\documentclass{article}
\usepackage{iclr2027_conference,times}
\iclrfinalcopy

\usepackage{amsmath,amssymb,amsfonts,mathtools,amsthm}
\usepackage{booktabs}
\usepackage{multirow}
\usepackage{graphicx}
\usepackage{xcolor}
\usepackage{listings}
\usepackage{hyperref}
\usepackage{url}
\hypersetup{
  hidelinks, pdftitle={Same Function, Different Gradients: Evaluating Gradient Reliability under Reparameterization}, pdfauthor={Bum Jun Kim}
}

\definecolor{grrcodebg}{HTML}{F7F9FC}
\definecolor{grrcodeframe}{HTML}{CBD5E1}
\definecolor{grrcodetext}{HTML}{1F2937}
\definecolor{grrcodekeyword}{HTML}{1D4ED8}
\definecolor{grrcodecomment}{HTML}{15803D}
\definecolor{grrcodestring}{HTML}{B42318}
\definecolor{grrcodefunction}{HTML}{0F766E}
\definecolor{grrcodetype}{HTML}{7C3AED}
\definecolor{grrcodeconstant}{HTML}{9A3412}
\definecolor{grrcodenumber}{HTML}{64748B}

\lstdefinestyle{grrpython}{
  language=Python, basicstyle=\ttfamily\scriptsize\color{grrcodetext}, keywordstyle=\color{grrcodekeyword}, morekeywords=[2]{True,False,None}, keywordstyle=[2]\color{grrcodeconstant}, commentstyle=\color{grrcodecomment}, stringstyle=\color{grrcodestring}, emph={minimal_grr_attention_audit,paired_scores,max_logit_delta, prediction_flip_rate,backward_on_batches, attention_head_gradient_scores, attention_head_path_normalized_gradient_scores,topk_indices, spearman,jaccard,clone_model,apply_qk_head_scaling, apply_vo_head_scaling}, emphstyle=\color{grrcodefunction}, emph={[2]Sequence,Literal,EncodedBatch,Module,device}, emphstyle={[2]\color{grrcodetype}}, backgroundcolor=\color{grrcodebg}, numbers=left, numberstyle=\tiny\color{grrcodenumber}, numbersep=7pt, columns=fullflexible, keepspaces=true, showstringspaces=false, breaklines=true, frame=single, frameround=tttt, rulecolor=\color{grrcodeframe}, framerule=0.5pt, framesep=4pt, xleftmargin=1.5em, framexleftmargin=1.2em, aboveskip=0.7em, belowskip=0.7em
}

\title{Same Function, Different Gradients:\\
Evaluating Gradient Reliability\\
under Reparameterization}
\author{Bum Jun Kim\thanks{Corresponding author}\\
The University of Tokyo\\
\texttt{bumjun.kim@weblab.t.u-tokyo.ac.jp}}
\date{}

\newcommand{\R}{\mathbb{R}}
\newcommand{\E}{\mathbb{E}}
\newcommand{\cD}{\mathcal{D}}
\newcommand{\cG}{\mathcal{G}}
\newcommand{\cL}{\mathcal{L}}
\newcommand{\norm}[1]{\lVert #1\rVert}
\newcommand{\Top}{\operatorname{Top}}

\newcommand{\Eqref}[1]{Eq.~\ref{#1}}
\newcommand{\Secref}[1]{Sec.~\ref{#1}}

\newtheorem{proposition}{Proposition}
\newtheorem{definition}{Definition}
\newtheorem{theorem}{Theorem}
\newtheorem{lemma}{Lemma}
\newtheorem{corollary}{Corollary}

\begin{document}
\maketitle
\lhead{}

\begin{abstract}
Gradients are no longer used only to train models: Gradient magnitudes are now measurement instruments, used to rank attention heads for pruning, to decide per-example clipping in private training, to allocate adapter rank budgets, and to debug models. Across these workflows, larger gradient magnitudes are taken to mean greater importance. Even for a fixed input--output function, these gradients are computed in one of many possible parameter coordinate systems. For example, attention scores depend on the query and key projections only through the paired product $W_QW_K^\top$. Scaling a head's query projection by $c$ and the corresponding key projection by $1/c$ therefore leaves every output logit unchanged. This change in parameter coordinates rescales the corresponding query and key gradient norms by $1/c$ and $c$, respectively. To test whether gradient-based decisions remain unchanged under these coordinate changes, we introduce Gradient Reliability under Reparameterization (GRR). GRR rewrites a trained model in new parameter coordinates that compute exactly the same function, verifies that predictions are unchanged, and then tests whether the decisions derived from the gradients are unchanged too. Our audits span GLUE classifiers, a ViT on CIFAR-10, GPT-2-medium and GPT-2-large, a 1.1B-parameter Llama-style model on WikiText-2, and six trained LoRA adapters. Empirically, predictions are preserved throughout, but both the gradients and the gradient-based decisions change. Across 66 GLUE attention audits, the median top-3 overlap of raw head-gradient rankings is only $1/3$, and rankings and pruning masks change even under the mildest rescalings we test. In the largest of 14 global-pruning audits, the shifted ranking selects head masks that differ in four of 22 pruned heads and reach $50.92\%$ and $94.84\%$ validation accuracy on one ELECTRA SST-2 model. We prove that same-function reparameterizations can realize any component ranking and make gradient scores arbitrarily large, so these gradient-based measurements are not properties of the learned function at all. Beyond diagnosis, we propose symmetry-aware remedies: Function-space controls and canonical balancing restore identical decisions in every audit we run, and a quotient-space score is provably invariant under any invertible change of basis in the low-rank factors. Gradient-based evidence must therefore be verified to remain unchanged under same-function reparameterization, or be independent of coordinate choice by definition; otherwise, the evidence cannot be trusted.
\end{abstract}

\section{Introduction}

Gradients are used for more than model training. Quantities derived from gradients have become general-purpose measurement tools, used for structured pruning \citep{michel2019,molchanov2017}, compression of large language models \citep{ma2023llmpruner}, differential privacy clipping \citep{abadi2016}, low-rank adaptation (LoRA) rank allocation \citep{zhang2023adalora}, and model debugging through training-example influence \citep{koh2017,grosse2023influence}. These workflows convert parameter, activation, or mask derivatives into scalar scores---often norms or Taylor sensitivities---that decide which components are pruned, which examples are clipped or flagged, and how rank budgets are allocated. These scores are treated as evidence of what matters to the represented input--output function.

\begin{figure}[t!]
\centering
\includegraphics[width=\linewidth]{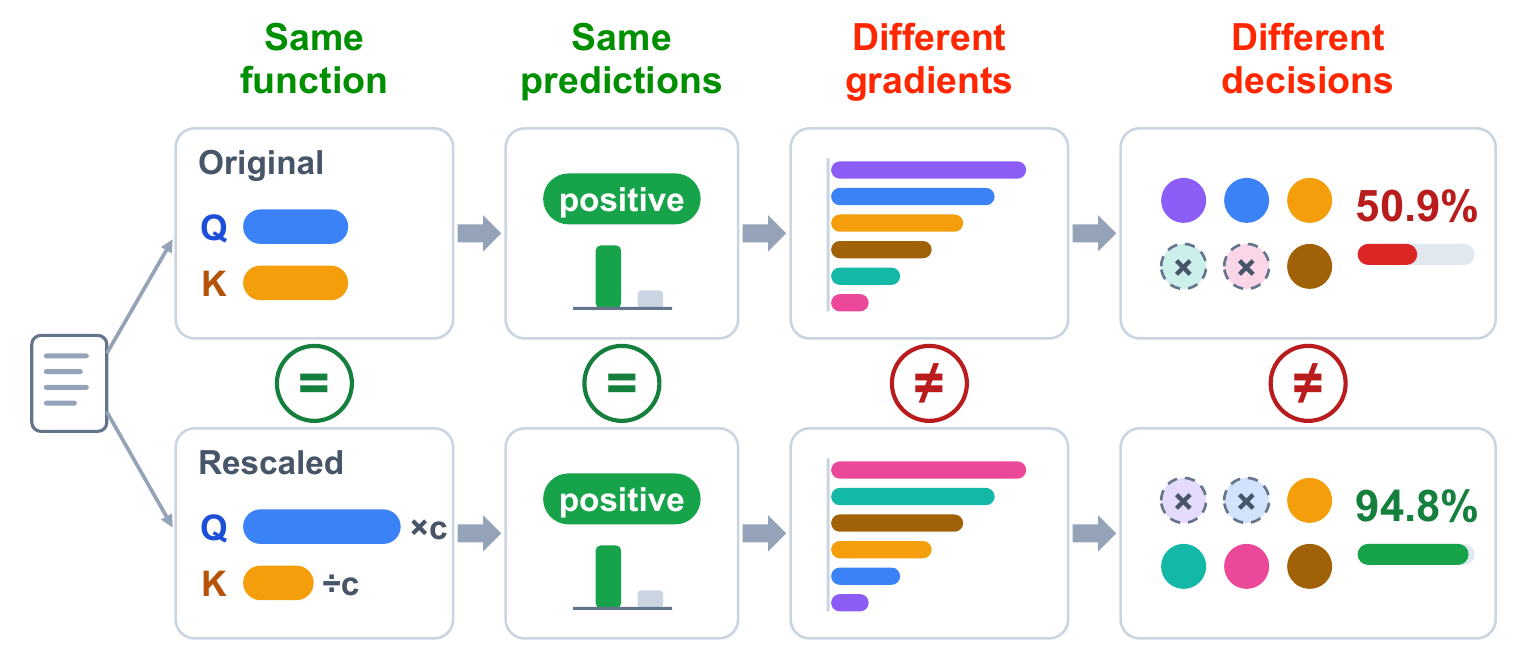}
\caption{Overview of the GRR audit. Scaling a head's query projection by $c$ and the corresponding key projection by $1/c$ preserves the model's function and predictions. This coordinate change rescales the query and key gradients by $1/c$ and $c$, respectively. Applying different scales across heads changes the raw-gradient head ranking. Gradient-guided pruning therefore keeps different heads, and the two pruned ELECTRA models differ substantially in SST-2 validation accuracy, as detailed in \Secref{sec:downstream}.}
\label{fig:overview}
\end{figure}

Yet each such score is computed in a particular coordinate system on the network's parameters. The choice matters because parameter gradients depend on the coordinates even when the input--output function is fixed. Raw gradient scores therefore reflect the coordinate choice as well as the represented function. Neural-network parameter spaces classically admit transformations that leave the input--output map unchanged \citep{hechtnielsen1990}. If a differentiable, locally invertible transformation $T:\Theta\to\Theta$ of the parameter space $\Theta$ satisfies $f_{T(\theta)}(x)=f_\theta(x)$ for every input $x\in\mathcal{X}$, then $T$ is a function-preserving reparameterization, and the differential of the loss is transported by the chain rule,
\begin{equation}
\label{eq:transport}
    \nabla_{\theta'}\cL = (DT^{-1}(\theta'))^\top\nabla_\theta\cL, \qquad \theta'\coloneqq T(\theta).
\end{equation}
Common gradient-based measurements are computed in a chosen Euclidean coordinate system. The resulting quantities and decisions, including gradient norms, coordinate-wise Taylor scores, diagonal Fisher estimates, clipping decisions, and top-$k$ rankings, can change with the coordinates. The function is unchanged by $T$, but the coordinate representation of the parameter gradient can change, as can scores computed from that representation.

An orbit is the set of parameter vectors reachable from one model through the chosen function-preserving transformations. These vectors represent the same input--output function in different parameter coordinates.

For example, Transformer attention admits such a function-preserving reparameterization. Attention scores depend on the query and key projections only through the paired product $W_QW_K^\top$, so replacing $(W_{Q,h},W_{K,h})$ by $(c_hW_{Q,h},c_h^{-1}W_{K,h})$ leaves every output logit unchanged while scaling the two head-gradient norms by $c_h^{-1}$ and $c_h$. The head still attends to the same tokens and contributes the same output. Only the units used to measure the head have changed, but this coordinate change rescales the head's parameter gradients. By choosing $c_h$, we can make the same head report an arbitrarily large gradient score without changing any prediction. The represented function therefore does not uniquely determine this raw gradient score---a failure of identifiability, not of measurement precision. We must therefore reassess the reliability of gradient-based evidence: Gradient norms, head rankings, clipping decisions, and rank budgets must be audited under coordinate changes that preserve the function. Input-side attributions are unaffected, since these attributions are determined by the function alone \citep{sundararajan2017}.

\paragraph{Same-function coordinate changes occur in practice.}
Data-free quantization equalizes weight ranges across adjacent layers, post-training quantization migrates activation scale into the paired weight matrix, and structured slicing rotates the internal representation before deleting rows and columns \citep{nagel2019dfq,xiao2023smoothquant,ashkboos2024slicegpt}. Each of these steps preserves the represented function and changes only how that function is written down. Compression pipelines routinely chain such a step with a gradient-based pruning or rank-allocation step, so the gradient evidence consumed downstream is read in a coordinate that an earlier and purely functional optimization happened to choose. No deliberate transformation is even required to produce scale imbalance. The eight public checkpoints we audit already have unequal paired projection norms: Balancing the paired projections within each checkpoint's orbit requires scales in $[0.837,1.077]$. The coordinate choice underlying gradient evidence is therefore already arbitrary.

To test whether gradient-based decisions remain unchanged under such coordinate changes, we introduce Gradient Reliability under Reparameterization (GRR): The audit holds the represented input--output function fixed, changes only the parameter coordinates, and tests whether the measurements read from the gradients---and the decisions those measurements drive---remain unchanged. We instantiate this invariance check with exact Transformer symmetries and verify forward preservation by checking that the two coordinates produce numerically identical outputs. We then examine gradient evidence and compare downstream decisions such as pruning masks, clipping labels, LoRA rank choices, and paired-product optimizer diagnostics. In the largest of our global GLUE pruning audits, two same-function coordinates of one ELECTRA SST-2 model make identical predictions, yet the pruning decisions differ sharply: The raw gradient ranking selects head masks that differ in four of the 22 pruned heads, and the two pruned models reach $50.92\%$ and $94.84\%$ validation accuracy. Figure~\ref{fig:overview} illustrates this audit. GRR holds the represented function fixed and tests whether an existing diagnostic and the corresponding downstream decision are well defined along exact symmetry orbits. We also propose symmetry-aware remedies. Function-space ablation, a path-normalized score, and canonical balancing make the two coordinates select identical decisions in every audit we run. A quotient-space score remains unchanged under any invertible change of basis in the low-rank factors. These remedies provide a drop-in replacement for a diagnostic that fails the audit.

\paragraph{Distinction from prior work.}
That Euclidean gradients depend on coordinates is classical \citep{amari1998,martens2020}, and prior work has used the freedom to change parameter coordinates in three ways: to question generalization claims such as flatness \citep{dinh2017}, to design scores that are invariant by construction \citep{neyshabur2015,tanaka2020}, and to exploit motion along symmetry orbits as an optimization resource \citep{kunin2021,zhao2022teleport}. Reliability tests for explanations are closest in spirit, but existing tests perturb the model \citep{adebayo2018} or the input \citep{kindermans2019} and test whether the explanation notices. GRR applies a complementary test: The audit leaves the model's behavior exactly intact, verified numerically before any gradient is computed, and tests whether a diagnostic that a pipeline already executes still returns the same decision. Correspondingly, every strict ordering of raw component scores is attainable on a single same-function orbit, and the attainable score ranges have no finite upper bound. To the best of our knowledge, the effects of these coordinate changes on downstream gradient evidence and decisions have not been systematically audited. Exact natural-gradient constructions are invariant in principle, but the diagonal approximations that pipelines actually use are among the measurements we audit. Appendix~\ref{app:related} details what GRR shares with, and where GRR departs from, natural-gradient geometry, quotient geometry, structured pruning, privacy, adapters, and explanation reliability.

\paragraph{Contributions.}
We formulate GRR as an operational audit of the identifiability of gradient-derived evidence, built from exact, architecture-native symmetries. Theoretically, we show that same-function reparameterizations can arbitrarily change raw component rankings, characterize the resulting ambiguity, and derive symmetry-aware alternatives, including a quotient-space score that is invariant to any invertible change of basis in the low-rank factors. Empirically, we trace these changes to pruning, clipping, optimizer, and LoRA decisions across encoder, vision, and decoder-only models.

\section{GRR Audit Framework}

Let $f_\theta:\mathcal{X}\to\mathcal{Y}$ be a neural network with parameters $\theta\in\Theta$, where $\Theta$ is the set of possible parameter vectors. Let $\ell(f_\theta(x),y)$ define a loss. A transformation $T:\Theta\to\Theta$ is function-preserving on $\mathcal{X}$ if
\begin{equation}
    f_{T(\theta)}(x)=f_\theta(x) \qquad \text{for all }x\in\mathcal{X}.
\end{equation}
On a finite audit set $\cD$, we check that the output logits $z_\theta(x)$ remain numerically unchanged using the maximum logit difference $\delta_f(T)$ defined in \Eqref{eq:forward-delta}. We also report the prediction flip rate: the fraction of audited predictions for which the highest-logit class or next token changes under $T$.

\paragraph{Audited gradients.}
Practitioners may compute gradients with respect to the input, an intermediate activation, a component mask, or a parameter block. The chain rule links these gradients, but each measures sensitivity to a different variable. DP-SGD clips per-example parameter gradients \citep{abadi2016}. LLM-Pruner uses weight derivatives to rank parameter groups for pruning \citep{ma2023llmpruner}, and AdaLoRA builds rank-importance scores from parameter--gradient products \citep{zhang2023adalora}. GRR therefore focuses on parameter gradients and on gradient-derived measurements and decisions.

Pruning also uses derivatives with respect to other variables. \citet{michel2019} score heads using derivatives with respect to head masks, while \citet{molchanov2017} score channels using activation--gradient products. These mask and Taylor scores differ from the raw parameter-gradient norms audited below. Weight--gradient and activation--gradient products cancel reciprocal scalar rescaling, so GRR tests the complete score and decision.

For a fixed input and label, the input gradient of the loss is unchanged by any $T$ with $f_{T(\theta)}=f_\theta$. Input gradients therefore remain unchanged under these transformations and are not the target of this audit. However, activation and parameter gradients can change because $T$ changes the coordinates used to compute these gradients. \Eqref{eq:transport} computes the parameter gradient in the new coordinates from the original gradient when $T$ is locally invertible at $\theta$ and the audited loss depends only on the represented function. These conditions hold for every symmetry used in this paper. The same reasoning applies to activation gradients when $T$ changes the coordinates of those activations.

Let $R(\theta,\cD)$ be a gradient-derived measurement, such as a layer-level gradient norm, head ranking, per-example gradient norm, clipping decision, internal-component attribution score, or rank-importance score. A function-level measurement should satisfy $R(\theta,\cD)\approx R(T(\theta),\cD)$ after correcting known label permutations and coordinate transports. A gradient reliability failure occurs when $f_\theta$ and $f_{T(\theta)}$ make the same predictions but $R$ changes enough to alter a downstream decision.

To audit such failures, we define a GRR report that records function preservation alongside changes in measurements and decisions.

\begin{definition}[GRR report]
\label{def:grr}
Given a model $\theta$, a function-preserving transformation $T$, an audit set $\cD$, a measurement $R$, and a downstream decision rule $D$, write $R_0\coloneqq R(\theta,\cD)$ and $R_T\coloneqq R(T(\theta),\cD)$ for the measurements before and after the transformation. The GRR report consists of
\par\noindent
\begin{minipage}[t]{0.54\linewidth}
\begin{equation}
\label{eq:forward-delta}
    \delta_f(T)\coloneqq \max_{x\in\cD}\norm{z_\theta(x)-z_{T(\theta)}(x)}_\infty,
\end{equation}
\end{minipage}\hfill
\begin{minipage}[t]{0.42\linewidth}
\begin{equation}
\label{eq:grr-correlation}
    \vphantom{\max_{x\in\cD}\norm{z_\theta(x)-z_{T(\theta)}(x)}_\infty} \rho_R\coloneqq\operatorname{Spearman}(R_0,R_T),
\end{equation}
\end{minipage}
\par\noindent
\begin{minipage}[t]{0.54\linewidth}
\begin{equation}
\label{eq:grr-overlap}
    O_k\coloneqq\frac{|\Top_k(R_0)\cap\Top_k(R_T)|}{k},
\end{equation}
\end{minipage}\hfill
\begin{minipage}[t]{0.42\linewidth}
\begin{equation}
\label{eq:grr-decision}
    \vphantom{\frac{|\Top_k(R_0)\cap\Top_k(R_T)|}{k}} \Delta_D\coloneqq d(D(R_0),D(R_T)).
\end{equation}
\end{minipage}
\par\noindent
Here $d$ denotes a task-specific comparison; later audits report set agreement for masks, metric gaps for downstream utility, and flip rates for binary clipping decisions. For binary clipping decisions $b_i\coloneqq\mathbf{1}\{\norm{g_i}_2>\tau\}$, $\Delta_D$ is the flip rate. Here $O_k$ is the top-$k$ overlap fraction. Unlike the mask comparisons reported later, this overlap fraction is not a Jaccard index.
\end{definition}

The audit proceeds as follows. First, we check forward preservation using $\delta_f(T)$ and prediction flips. Only after the forward-preservation check passes do we compare rankings and score scales, then pruning, clipping, or rank-selection decisions. Finally, we repeat the comparison with a proposed remedy under the same verified transformation. If the forward check fails, the audit is invalid: We cannot attribute later differences to a change of coordinates alone. If the forward check passes but scores or decisions change beyond numerical tolerance, the diagnostic depends on the parameter coordinates even though the function is unchanged. If the remedy's scores or decisions also change between the two coordinates beyond numerical tolerance, the remedy has not removed that dependence. Appendix~\ref{app:minimal-grr-code} gives a compact executable reference implementation of this audit sequence for an attention-pair audit.

\section{Function-Preserving Symmetries and Remedies}

\subsection{Bilinear Gauge}

To see how changing parameter coordinates can change gradient scores while keeping the function fixed, consider a bilinear product $M\coloneqq BA$. Here, a gauge transformation is a change of the factors that leaves the product $BA$ unchanged. For any invertible $G\in GL(r)$,
\begin{equation}
    A'\coloneqq G^{-1}A,\qquad B'\coloneqq BG
\end{equation}
leaves $BA$ unchanged. As $G$ varies, these pairs form the orbit of $(B,A)$. If $\cL$ depends on $(A,B)$ only through $BA$, then the gradients transform as
\begin{equation}
    \nabla_{A'}\cL = G^\top\nabla_A\cL, \qquad \nabla_{B'}\cL = \nabla_B\cL G^{-\top}.
\end{equation}
For scalar $G=cI$ with $c>0$,
\begin{equation}
    \norm{\nabla_{A'}\cL}_F=c\norm{\nabla_A\cL}_F,\qquad \norm{\nabla_{B'}\cL}_F=c^{-1}\norm{\nabla_B\cL}_F.
\end{equation}
Thus factor-level gradient norms and rank-coordinate scores can be rescaled without changing the represented function. The scalar-gauge results only require this reciprocal scaling to be an exact symmetry. Literal dependence on a single product $BA$ is sufficient but not necessary.

For a factor pair, the squared joint gradient score follows $S(c)^2=ac^2+bc^{-2}$, where $a\coloneqq\norm{\nabla_A\cL}_F^2$ and $b\coloneqq\norm{\nabla_B\cL}_F^2$. For $a,b>0$, the score has no finite upper bound. Lemma~\ref{lem:scalar-score-range} in Appendix~\ref{app:scalar-results} gives the exact range of the joint gradient score.

\begin{theorem}[Any raw component ranking at a fixed function]
\label{thm:arbitrary-ranking}
Let $\cL$ be differentiable, let $H\geq2$, and suppose that, for every choice of positive scales $c_1,\ldots,c_H$, the independent reciprocal action $A'_h\coloneqq c_h^{-1}A_h$, $B'_h\coloneqq c_hB_h$ with $c_h>0$ is an exact function-preserving reparameterization and leaves $\cL$ invariant. At a fixed representative, assume both factor gradients are nonzero for every $h$, and define the joint score by $S_h(c_h)^2\coloneqq\norm{\nabla_{A'_h}\cL}_F^2+\norm{\nabla_{B'_h}\cL}_F^2$ with $S_h(c_h)\geq0$. For every permutation $\pi$ of $\{1,\ldots,H\}$, there are same-function scales $c_1,\ldots,c_H$ such that
\begin{equation}
    S_{\pi(1)}(c_{\pi(1)}) <\cdots< S_{\pi(H)}(c_{\pi(H)}).
\end{equation}
\end{theorem}

Thus any raw top-$k$ or bottom-$k$ set is attainable. Appendix~\ref{app:scalar-results} gives the corollary, proofs, and checkpoint construction.

Under the stated nonzero-gradient conditions, the theorem applies directly to independently scaled standard attention heads, SwiGLU up--down channels, and diagonal LoRA rank coordinates. In GQA, the same conclusion applies at the level of independently gaugeable shared groups rather than individual query heads.

The same dependence reaches Euclidean factor-gradient steps: At scalar gauge $c$, the first-order product displacement is $-\eta(c^2U+c^{-2}V)$ for $U\coloneqq B\nabla_A\cL$ and $V\coloneqq\nabla_B\cL A$. Appendix~\ref{app:proofs} gives the exact finite-step and direction results. We call $U+V$, with the common factor $-\eta$ omitted, the paired-product tangent; classifier-factor and LoRA diagnostics use the full effective product.

The same coordinate dependence can affect clipping decisions and diagonal approximations. Raw gradient magnitudes, component selections, and update measurements therefore require checks under exact function-preserving coordinate changes before serving as evidence about the learned function.

\subsection{Transformer Attention and MLP Symmetries}

In row-vector notation, multi-head attention \citep{vaswani2017} computes
\begin{equation}
    Q\coloneqq XW_Q+b_Q,\qquad K\coloneqq XW_K+b_K,\qquad S\coloneqq\frac{QK^\top}{\sqrt{d_h}}+M_{\mathrm{mask}}.
\end{equation}
For a head-wise scalar gauge $c_h>0$,
\begin{equation}
    W_{Q,h}'\coloneqq c_h W_{Q,h},\quad b_{Q,h}'\coloneqq c_h b_{Q,h},\quad W_{K,h}'\coloneqq c_h^{-1}W_{K,h},\quad b_{K,h}'\coloneqq c_h^{-1}b_{K,h}
\end{equation}
preserves $Q_hK_h^\top$ exactly in real arithmetic. The Q--K gradients scale oppositely, so head-wise gradient norms and rankings need not be preserved.

The V--O symmetry is analogous. For a head contribution $H_h\coloneqq P_h(XW_{V,h}+b_{V,h})W_{O,h}$, the head-wise reciprocal scaling
\begin{equation}
    W_{V,h}'\coloneqq c_hW_{V,h},\qquad b_{V,h}'\coloneqq c_hb_{V,h},\qquad W_{O,h}'\coloneqq c_h^{-1}W_{O,h}
\end{equation}
leaves $H_h$ unchanged. This V--O symmetry is independent of positional encoding details.

In models that use rotary position embedding (RoPE) \citep{su2024roformer}, full Q--K rotations are restricted, but scalar head scaling still commutes with RoPE. GQA \citep{ainslie2023gqa} retains group-wise gauges, and Llama-style SwiGLU blocks \citep{touvron2023llama} admit $W'_{u,j}=c_jW_{u,j}$ and $W'_{d,j}=c_j^{-1}W_{d,j}$ with the gate fixed. Appendix~\ref{app:proofs} gives the exact architecture details. Appendix~\ref{app:related} covers deployed same-function coordinate moves.

\subsection{LoRA and Low-Rank Adapters}

LoRA \citep{hu2022} uses an effective update $\Delta W\coloneqq(\alpha/r)BA$. For any $G\in GL(r)$, the gauge $B'\coloneqq BG$, $A'\coloneqq G^{-1}A$ preserves $\Delta W$. Because the fixed prefactor $\alpha/r$ can be absorbed into either factor, we suppress the prefactor in the theoretical statements below and write $\Delta W\coloneqq BA$. The prefactor value is a design choice that affects how LoRA trains at a given rank \citep{kalajdzievski2023rslora}, but the prefactor is identical in the two coordinates that an audit compares, so the prefactor does not affect the gauge argument.

Diagonal monomial gauges preserve rank components up to rescaling and relabeling, whereas non-diagonal $GL(r)$ gauges rotate the basis. Under the conditions of Theorem~\ref{thm:rank-ablation-orbit}, deleting one labeled rank coordinate can produce a product change with any Frobenius norm in $[\sigma_r(BA),\infty)$, while the product before deletion stays fixed. Appendix~\ref{app:proofs} gives the proof. Consequently, factor-gradient norms and rank-coordinate importance can depend on an arbitrary basis in the low-rank space.

\subsection{Symmetry-Aware Remedies}
\label{sec:remedies}

The goal is to keep the measurement and the decision derived from that measurement unchanged when only the parameter coordinates change. We pursue this goal by balancing paired factors before measuring, canceling scale factors within the score, or measuring the represented function directly.

\paragraph{Gauge-invariant product-manifold score.}
We construct a first-order score on the fixed-rank product manifold $\mathcal{M}_r\coloneqq\{M\in\R^{m\times n}:\operatorname{rank}(M)=r\}$. Let $B\in\R^{m\times r}$ have full column rank and $A\in\R^{r\times n}$ have full row rank. Let $\overline{\cL}$ be continuously differentiable in a neighborhood of $M\coloneqq BA\in\mathcal{M}_r$, let $\cL(B,A)\coloneqq\overline{\cL}(BA)$, and set $D_A\coloneqq\nabla_A\cL$, $D_B\coloneqq\nabla_B\cL$, $P\coloneqq B^\top B$, and $Q\coloneqq AA^\top$, with $\Psi\coloneqq D_AA^\top=B^\top D_B$. Define the nonnegative score $\mathcal{I}_{\mathrm{quot}}(B,A)$ by
\begin{equation}
\label{eq:quotient-gradient-score}
    \mathcal{I}_{\mathrm{quot}}^2(B,A) \coloneqq\operatorname{tr}(P^{-1}D_AD_A^\top) +\operatorname{tr}(Q^{-1}D_B^\top D_B) -\operatorname{tr}(Q^{-1}\Psi^\top P^{-1}\Psi).
\end{equation}

\begin{theorem}[Product-manifold gradient from factor gradients]
\label{thm:quotient-gradient}
With the notation above, equip $\mathcal{M}_r$ with the ambient Frobenius metric. If $Z\coloneqq\nabla_M\overline{\cL}(M)$, $\Pi_B\coloneqq B P^{-1}B^\top$, and $\Pi_A\coloneqq A^\top Q^{-1}A$, then
\begin{equation}
\label{eq:quotient-gradient-projection}
    \mathcal{I}_{\mathrm{quot}}(B,A) =\norm{\Pi_BZ+Z\Pi_A-\Pi_BZ\Pi_A}_F =\norm{\operatorname{grad}_{\mathcal{M}_r}\overline{\cL}(M)}_F.
\end{equation}
\end{theorem}

\begin{corollary}[Full-gauge invariance and stationarity]
\label{cor:quotient-gradient}
Under the conditions of Theorem~\ref{thm:quotient-gradient}, the score in \Eqref{eq:quotient-gradient-score} is exactly invariant under every transformation $(B,A)\mapsto(BG,G^{-1}A)$ with $G\in GL(r)$. The score vanishes if and only if $D_A=D_B=0$.
\end{corollary}

The score requires only two $r\times r$ Gram solves and factor-gradient contractions. The score does not assign importance to individual rank coordinates, which are not objects on the quotient. Proofs appear in Appendix~\ref{app:full-gl-proofs}; further interpretation and the full-gauge gradient envelope appear in Appendix~\ref{sec:full-gl-remedies}. Appendix~\ref{app:remedy-details} defines the empirical-Fisher and function-space controls.

\paragraph{Canonical balancing.}
For scalar gauges and $M\coloneqq BA$ with $\norm{A}_F\norm{B}_F>0$, setting $c\coloneqq\sqrt{\norm{A}_F/\norm{B}_F}$ in $(B',A')\coloneqq(cB,c^{-1}A)$ balances the paired norms. Appendix~\ref{app:remedy-details} gives the full balancing construction, the proof of orbit invariance, and implementation details.

\paragraph{Path-normalized gradient score.}
For scalar gauges, a computationally inexpensive normalization is to score a parameter block by the product of the parameter norm and the gradient norm of that block, vectorizing matrix-valued blocks before taking the Euclidean norm:
\begin{equation}
    I_h^{\mathrm{path}} \coloneqq \left( \sum_{p\in I_h} \norm{\theta_p}_2^2 \norm{\nabla_{\theta_p}\cL}_2^2 \right)^{1/2}.
\end{equation}
Each product is invariant under a reciprocal scalar gauge of a parameter block and the gradient of that block, as verified in Appendix~\ref{app:proofs}.

\section{Experiments}

\subsection{Setup}

We use GRR to test whether changing parameter coordinates alters gradient-based measurements and decisions while the model's function stays fixed. For exact Q--K and V--O rescaling, our attention audits cover 66 layer-level GLUE audits, a ViT on CIFAR-10, and GPT-2-medium, GPT-2-large, and TinyLlama on WikiText-2. We also audit TinyLlama's SwiGLU up--down rescaling on WikiText-2.

For low-rank basis dependence, we apply GRR to exact low-rank factorizations of the seven GLUE classifier heads and to six independently trained LoRA adapters: three rank-16 DistilBERT adapters on SST-2 and three rank-8 RoBERTa adapters on MRPC.

The main GLUE gradient-score audits use 128 validation examples, the clipping audits use 48 examples, and downstream accuracy is evaluated on the full validation splits. For a nonnegative score vector $s$, the mean ratio denotes $\operatorname{mean}(s_{\mathrm{transformed}})/\operatorname{mean}(s_{\mathrm{original}})$. All reported accuracy and perplexity gaps are absolute differences between decisions induced by the two same-function coordinates.

Every comparison first verifies forward preservation. All classification audits have zero label flips. Decoder-only audits have median next-token argmax flip rate zero; the only nonzero case is a finite-precision GPT-2-large layer-9 Q--K audit with a $0.0492\%$ flip rate. Appendix~\ref{app:additional-experiments} gives the complete model identifiers, audit budgets, gauge profiles, training recipes, baseline results in Table~\ref{tab:baseline}, and preservation checks in Table~\ref{tab:forward}.

\subsection{Attention-Head Gradient Reliability}

Table~\ref{tab:aggregate} aggregates 66 layer-level attention audits. Raw Euclidean head-gradient norms are unstable: The median top-3 overlap is $1/3$, and some audits have zero overlap. In the median audit, changing coordinates replaces two of the three heads that raw gradients call most important, even though the model makes the same predictions.

Mean gradient scale changes range from $1.98\times$ to $3.52\times$. The paired-product tangent of a raw Euclidean parameter-gradient step changes even more, with ratios from $9.97\times$ to $23.12\times$. The same model and learning rate can therefore imply a very different first-order change in the paired product. Per-example clipping decisions, using the original median norm as threshold, flip for a median $20.8\%$ of examples and up to $50.0\%$. Parameter scaling alone changes which per-example gradients the clipping rule reduces.

Function-space head ablation is stable in every audit, with top-3 overlap $1.0$ in all 66 audits. The path-normalized gradient score is also stable for these scalar gauges, with top-3 overlap $1.0$ and mean ratio between $0.999997$ and $1.000003$. Canonically balanced parameter gradients recover stable decisions as well, with top-3 overlap $1.0$ in all 66 audits. These controls make head selection independent of the starting parameter scales under the tested scalar symmetry. Figure~\ref{fig:summary}(a) in Appendix~\ref{app:additional-experiments} compares raw and symmetry-aware scores across the same 66 audits.

\begin{table}[t!]
\centering
\caption{Layer-level attention GRR over 66 audits. Agreement denotes top-3 overlap except for per-example clipping, where agreement denotes the decision flip rate. Values are medians with ranges in parentheses. Raw measurements are unstable; the controls are stable under the scalar gauges tested.}
\label{tab:aggregate}
\small
\begin{tabular}{lcc}
\toprule
Measurement & Agreement or flip rate & Mean ratio \\
\midrule
Raw head-gradient norm & $0.333$ $(0.000,1.000)$ & $2.98$ $(1.98,3.52)$ \\
Raw paired-product tangent & $0.333$ $(0.000,0.667)$ & $16.84$ $(9.97,23.12)$ \\
Per-example clipping & $20.8\%$ $(6.25,50.0)$ & Not applicable \\
\midrule
Function-space ablation & $1.000$ $(1.000,1.000)$ & $\approx 1$ \\
Path-normalized gradient & $1.000$ $(1.000,1.000)$ & $1.00$ $(1.00,1.00)$ \\
Canonically balanced gradient & $1.000$ $(1.000,1.000)$ & $\approx 1$ \\
\bottomrule
\end{tabular}
\end{table}

\begin{table}[t!]
\centering
\caption{GLUE decision-level effects; pp denotes percentage points. The ViT global-profile counterpart is reported in Table~\ref{tab:vit_global_profile}.}
\label{tab:decisions}
\small
\begin{tabular}{lccc}
\toprule
Decision rule & Jaccard median & Jaccard range & Max accuracy gap (pp) \\
\midrule
Single-layer raw bottom-3 pruning & 0.20 & $(0.00,1.00)$ & 1.47 \\
Single-layer raw top-3 ablation & 0.20 & $(0.00,1.00)$ & 10.05 \\
Single-layer function-space top-3 & 1.00 & $(1.00,1.00)$ & 0.00 \\
\midrule
Global raw bottom-15\% pruning & 0.517 & $(0.333,0.692)$ & 43.92 \\
Global raw top-15\% ablation & 0.294 & $(0.158,0.467)$ & 27.18 \\
Global canonically balanced bottom-15\% & 1.00 & $(1.00,1.00)$ & 0.00 \\
\bottomrule
\end{tabular}
\end{table}

\paragraph{Robustness and breadth.}
The failure persists across subsets, seeds, gauge strengths, architectures, and modalities. Over 324 DistilBERT SST-2 audits, mild gauges give mean raw top-3 overlap $0.787$, aggressive gauges give $0.512$, and all runs have zero prediction flips. Figure~\ref{fig:summary}(c) and Figure~\ref{fig:summary}(d) in Appendix~\ref{app:additional-experiments} show the smooth scale trend. Even the mildest tested rescaling changes head choices, and stronger rescaling increases the disagreement. Public checkpoints contain milder but nonzero paired-projection imbalance, and the balancing scales observed across the encoder and ViT checkpoints lie in $[0.837,1.077]$. The cross-domain audits show the same pattern: ViT raw top-3 overlap has median $1/3$, GPT-2-medium and GPT-2-large have median $1/3$, TinyLlama paired-product tangent overlap has median $0$, and the TinyLlama SwiGLU channel-gradient overlap has median $0.352$. Appendix~\ref{app:additional-experiments} reports the full robustness, checkpoint-scale, ViT, decoder-only, and SwiGLU results.

\subsection{Downstream Head Decisions}
\label{sec:downstream}

For each layer, we evaluate two decision rules: bottom-3 pruning, which removes heads judged least important, and top-3 ablation, which removes heads judged most important. Table~\ref{tab:decisions} summarizes these single-layer decisions and the global decisions below. Head sets selected for top-3 ablation using raw gradients have median Jaccard $0.2$ between the original and reparameterized coordinates, with minimum $0$. The maximum validation-accuracy gap between the ablations is $10.05$ percentage points. Raw bottom-3 pruning has the same median Jaccard and yields a gap of $1.47$ percentage points. These ranking disagreements determine which heads are actually removed and can change the accuracy of the resulting model. Function-space and canonically balanced top-3 decisions have Jaccard $1.0$ in every audit.

We also run a global downstream test on the GLUE checkpoints. We apply a head-wise gauge to every attention layer, compute one global raw-gradient ranking over all heads, and prune or ablate the bottom or top 15\% of heads. Raw global top-head ablation has median Jaccard $0.294$ and yields a validation-accuracy gap of up to $27.18$ percentage points. Raw global bottom-head pruning has median Jaccard $0.517$ and yields a gap of up to $43.92$ percentage points. In the largest case, an ELECTRA Q--K audit on SST-2, the original-coordinate mask gives $50.92\%$ accuracy while the reparameterized-coordinate mask gives $94.84\%$. Figure~\ref{fig:qual-global-mask} in Appendix~\ref{sec:global-head-case-study} localizes the four head substitutions behind this accuracy gap from global pruning. With identical predictions before pruning, coordinate choice alone separates accuracy close to the baseline from accuracy near chance afterward. Canonical balancing makes global bottom pruning and top ablation select identical masks, with zero validation-accuracy gap in all 14 GLUE global audits. Balancing fixes the disagreement, but the shared mask can still harm accuracy. Table~\ref{tab:vit_global_profile} reports the corresponding full-test vision result.

\paragraph{Mild-scale downstream calibration.}
Mild V--O gauges already change global masks in downstream evaluations. On MRPC, mild bottom-15\% pruning gives mask Jaccard $0.630$ and an accuracy gap of $17.89$ percentage points; on QNLI, mild top-15\% ablation gives Jaccard $0.419$ and a gap of $18.47$ percentage points. Large accuracy gaps can arise even under mild rescaling. The corresponding SST-2 accuracy gaps remain below one percentage point despite changed masks. Similar accuracy can hide disagreement about which heads matter. Appendix~\ref{app:additional-experiments} reports all mild, medium, and aggressive profile results.

\paragraph{Additional workflow audits.}
The same coordinate dependence propagates through iterative pruning, clipped updates, and optimizer-state diagnostics. In the iterative RoBERTa MRPC audit, the final raw top-ablation masks have Jaccard $0.244$, and the final accuracies differ by $13.48$ percentage points; canonical balancing restores identical schedules. Recomputing gradients at each round carries the coordinate dependence through the pruning schedule.

Diagonal empirical-Fisher normalization recovers top-3 overlap $1.0$ in the targeted clipping audit, whereas a transported AdamW state still yields a coordinate-dependent paired-product update. Transferring the optimizer's history correctly is therefore insufficient to make the update independent of parameter scaling. Appendix~\ref{app:additional-experiments} gives the complete iterative, Fisher-normalized, clipped-update, and AdamW analyses.

The decoder-only audits in Appendix~\ref{sec:decoder-only}, including the token-level analysis in Figure~\ref{fig:qual-lm-tokens}, also select different heads for the same unpruned language model, producing different next-token losses and perplexities.

\paragraph{Low-rank and adapter audits.}
Exact classifier-factor gauges and six trained LoRA adapters exhibit the same coordinate dependence: Factor-coordinate rankings, clipping decisions, and adapter-compression masks can change while logits are preserved. The raw scores can label different rank coordinates as expendable even when the adapter's predictions are identical. Appendix~\ref{sec:low-rank-factor-audits} reports the complete low-rank and trained-adapter results, including the symmetry-aware controls.

\section{Discussion}

Same-function coordinate changes can alter pruning, clipping, and update decisions. Appendix~\ref{app:workflow-implications} discusses these workflow implications. The failure concerns raw Euclidean measurements and diagnostics that retain coordinate dependence; fixed-function input gradients and invariant scores remain outside the scope of the coordinate-dependence claim. Invariance removes coordinate ambiguity but does not guarantee useful component selection. Appendix~\ref{app:checklist} gives the application procedure and acceptance criteria.

\section{Conclusion}

Gradient-derived measurements need not be properties of learned functions. GRR makes this distinction operational: The audit rewrites a trained model in two coordinates that provably compute the same function, verifies forward preservation, and tests whether gradient-derived measurements---and the decisions those measurements drive---change. Across Transformer workflows involving encoder and vision classifiers, decoder-only language models, and LoRA adapters, raw Euclidean gradients often change rankings, masks, clipping decisions, and paired-product updates, whereas function-space controls and symmetry-aware normalizations are stable under the scalar gauges tested. Our analysis shows that same-function orbits can realize any raw ranking and unbounded factor-gradient magnitudes, and yields a full-$GL(r)$-invariant quotient score. Gradient-based conclusions should therefore be audited under exact same-function reparameterizations or defined directly on quotient-space objects; the audit reduces to a checklist that a practitioner can apply to any new gradient-based diagnostic.

\clearpage
\section*{AI use statement }
Generative AI tools were used as assistive tools for writing and editing, mathematical proofreading and consistency checking, and code development and debugging. Any AI-assisted outputs incorporated into this work were reviewed and validated by the authors as appropriate. The authors take full responsibility for the final content of this work, including any text, claims, or artifacts produced with the assistance of generative AI.

\subsection*{Reproducibility Statement}
The theoretical assumptions and complete proofs appear in Appendix~\ref{app:proofs}. Experimental configurations, model and dataset identifiers, audit budgets, and full additional results appear in Appendix~\ref{app:additional-experiments}. Appendix~\ref{app:repro} documents artifacts, validation, result traceability, data construction, seeds, training recipes, and hardware configuration.

\bibliographystyle{iclr2027_conference}
\phantomsection
\addcontentsline{toc}{section}{References}
\bibliography{gra}

\clearpage
\appendix

\section{Notation}
\label{app:notation}

Table~\ref{tab:notation} collects the symbols used in the main text and in the proofs.

\begin{table}[h]
\centering
\caption{List of notation, grouped by the role each symbol plays in the audit.}
\label{tab:notation}
\small
\begin{tabular}{ll}
\toprule
Symbol & Meaning \\
\midrule
\multicolumn{2}{l}{\textbf{Model, loss, and audit}} \\
$f_\theta$ & Network with parameters $\theta$ \\
$\theta,\Theta$ & Parameter vector and parameter space \\
$\mathcal{X},\cD$ & Input space and finite audit set \\
$z_\theta(x)$ & Logits of $f_\theta$ at input $x$ \\
$\ell,\cL$ & Per-example loss and audited loss \\
$\eta$ & Learning rate of one gradient step \\
$T$ & Function-preserving reparameterization, $\theta'\coloneqq T(\theta)$ \\
$\delta_f(T)$ & Maximum logit difference induced by $T$ \\
$R(\theta,\cD)$ & Gradient-derived measurement under audit \\
$R_0,R_T$ & Original and transformed measurements in the GRR report \\
$\rho_R,O_k$ & Spearman correlation and top-$k$ overlap of $R$ \\
$D,\Delta_D$ & Downstream decision rule and decision change \\
$g_i,\tau,b_i$ & Per-example gradient, clipping threshold, clipping decision \\
\midrule
\multicolumn{2}{l}{\textbf{Symmetry group and quotient}} \\
$\cG,\phi_g$ & Symmetry group and group action on $\theta$ \\
$\mathcal{O}_\theta$ & Same-function orbit of $\theta$ \\
$\Theta/\cG$ & Quotient of same-function coordinates \\
\midrule
\multicolumn{2}{l}{\textbf{Bilinear factors and gauges}} \\
$A,B,M$ & Factors and factor product $M\coloneqq BA$ \\
$r$ & Rank of the bilinear factorization \\
$G$ & Gauge in $GL(r)$ acting by $(B,A)\mapsto(BG,G^{-1}A)$ \\
$c,c_h$ & Positive scalar gauge, per component $h$ \\
$D_A,D_B$ & Factor gradients $\nabla_A\cL$ and $\nabla_B\cL$ \\
$S(c),\Gamma(G)$ & Joint factor-gradient norm under scalar and full gauge \\
$U,V$ & Terms $B\nabla_A\cL$ and $\nabla_B\cL A$ in the paired-product tangent \\
$\delta M_c$ & Product change of one gradient step at gauge $c$ \\
$E_i,C_i(G)$ & $e_ie_i^\top$ and rank component $BGE_iG^{-1}A$ \\
$M_{-i}(G)$ & Product after deleting rank coordinate $i$ \\
$\sigma_r(M)$ & Smallest nonzero singular value of $M$ \\
$\norm{\cdot}_F,\norm{\cdot}_*$ & Frobenius and nuclear norms \\
$\mathcal{F}_r,\mathcal{M}_r$ & Full-rank factor pairs and rank-$r$ matrices \\
$P,Q,\Psi$ & Gram matrices $B^\top B$, $AA^\top$, and $D_AA^\top=B^\top D_B$ \\
$Z,\Pi_B,\Pi_A$ & Product gradient $\nabla_M\overline{\cL}$ and range projectors \\
$\mathcal{I}_{\mathrm{quot}}$ & Gauge-invariant product-manifold gradient norm \\
\midrule
\multicolumn{2}{l}{\textbf{Transformer components}} \\
$W_Q,W_K,W_V,W_O$ & Attention projections, with biases $b_Q,b_K,b_V$ \\
$d_h,H$ & Head dimension and number of audited components \\
$P_h,H_h$ & Attention probabilities and output of head $h$ \\
$c_g$ & Shared gauge of a GQA group $g$ \\
$W_g,W_u,W_d$ & SwiGLU gate, up, and down projections, channel gauge $c_j$ \\
$\Delta W,\alpha$ & LoRA update $(\alpha/r)BA$ and LoRA scaling factor \\
\midrule
\multicolumn{2}{l}{\textbf{Scores and remedies}} \\
$I_h,I_h^{\mathrm{path}}$ & Coordinates of head $h$ and path-normalized head score \\
$\widehat F_i,I_h^{\mathrm{Fisher}}$ & Diagonal empirical Fisher and Fisher-normalized score \\
$I_j^{\mathrm{func}}$ & Function-space ablation score of component $j$ \\
$\theta\setminus j$ & Parameters with the output path of $j$ removed \\
$r_h,s_h$ & Paired norm imbalance ratio and canonical balancing scale \\
\bottomrule
\end{tabular}
\end{table}

\section{Related Work and Distinction}
\label{app:related}

GRR operationalizes the classical coordinate dependence of Euclidean gradients as a decision-facing reliability audit: Fix the learned input--output function, change only the internal coordinate system, verify forward preservation numerically, and test whether a concrete conclusion survives on trained models and downstream decisions. This framing makes GRR closer to a measurement reliability test than to an optimization method.

\subsection{Reparameterization and Natural-Gradient Geometry}

Reparameterization dependence of Euclidean gradients is classical. \citet{amari1998} developed natural gradient methods to account for information geometry, \citet{martens2015} make this principle computationally useful for neural networks through Kronecker-factored approximate curvature and related approximations, and \citet{martens2020} analyzes the sense in which the resulting updates are invariant to smooth reparameterization. The same freedom destabilizes other parameter-space quantities: Sharpness measures and flat-minima claims can be sensitive to reparameterization \citep{dinh2017}, loss-surface visualizations require filter-normalized directions before the visualizations are compared \citep{li2018losslandscape}, and scale-invariant networks, path norms, and product norms show that parameter coordinates are not always meaningful \citep{neyshabur2015}. Normalization layers make the same freedom concrete: A weight block feeding a batch-normalized \citep{ioffe2015} or weight-normalized \citep{salimans2016} unit is positively scale-invariant, so rescaling that block leaves the function untouched while the corresponding gradient norm scales reciprocally \citep{vanlaarhoven2017}. The Q--K, V--O, and SwiGLU up--down gauges audited here are the paired-projection analogue of that freedom inside a Transformer block.

\subsection{Network Symmetry and Quotient Manifolds}

Symmetry and gauge freedom in neural networks supply the structural source of these equivalence classes \citep{hechtnielsen1990}, and recent surveys emphasize that large same-function parameter orbits are now a central object in overparameterized network theory \citep{zhao2026symmetry}. \citet{godfrey2022} catalogue the symmetry groups induced by common activation functions and the action of those groups on internal representations, which is the structural reason one learned function admits many coordinate descriptions. \citet{kunin2021} show that translation, scale, and rescale symmetries induce exact conservation laws in gradient-flow dynamics, complementary evidence that motion along symmetry directions carries no functional information, whereas \citet{zhao2022teleport} teleport parameters along a loss level set precisely because that move changes the local gradient. \citet{lim2024symmetries} instead reduce or remove parameter symmetries on the architecture side, while GRR holds the architecture and the learned function fixed and moves along the orbit to test whether a measurement is identifiable.

Discrete permutation symmetries of hidden units have been used in a different direction, to explain linear mode connectivity between independently trained solutions \citep{entezari2022} and to align and merge such solutions in weight space \citep{ainsworth2023rebasin}. Riemannian quotient geometry for full-rank bilinear factorizations is likewise standard in fixed-rank matrix optimization \citep{mishra2014,absil2008}. Our use of that geometry is measurement-facing, since Theorem~\ref{thm:gradient-orbit-envelope}, Corollary~\ref{cor:clipping-orbit}, Theorem~\ref{thm:rank-ablation-orbit}, and Corollary~\ref{cor:rank-ablation-deletion} turn factorization freedom into exact attainable sets with clipping and deletion consequences, and Corollary~\ref{cor:quotient-gradient} records an invariant fixed-rank product-manifold gradient score.

\subsection{Same-Function Coordinates in Deployed Transformer Pipelines}

Moving a Transformer along such an orbit is not a hypothetical operation. Data-free quantization equalizes weight ranges across adjacent layers through the scale equivariance of the activation function, which changes coordinates without changing the represented function \citep{nagel2019dfq}. Post-training quantization of Transformers applies a mathematically equivalent per-channel rescaling that migrates activation scale into the paired weight matrix \citep{xiao2023smoothquant}, and \citet{ashkboos2024slicegpt} use a computational invariance of Transformer blocks to rotate the representation before deleting rows and columns. These pipelines choose a new same-function coordinate for efficiency reasons. GRR tests whether gradient-derived evidence remains stable under such coordinate changes.

\subsection{Gradient-Based Pruning and Importance}

GRR complements work on structured pruning and head importance. \citet{michel2019} study which Transformer heads can be removed without hurting performance, and \citet{voita2019} find that a small number of specialized heads carries most of the work. Derivative-based importance has a long history in pruning, beginning with the second-order saliency of \citet{lecun1990} and continuing through first-order gradient and Taylor scores for weights, channels, and structures \citep{lee2019,molchanov2017,molchanov2019} and jointly learned coupled-structure masks \citep{xia2022}.

The same kind of evidence now drives structured compression of large language models: \citet{ma2023llmpruner} identify structurally coupled parameter groups and rank the groups using gradient-based importance, and \citet{frantar2023sparsegpt} prune in one shot through layer-wise second-order reconstruction, whereas \citet{sun2024wanda} deliberately replace gradients by a weight-magnitude and input-activation product. Closest to our concern on the score side, \citet{tanaka2020} derive conservation laws for synaptic saliency and design a synaptic-flow score whose invariance properties avoid layer collapse. That line designs a score with an invariance in mind. GRR instead tests whether an already-deployed score is identifiable from the learned function. Our one-shot, global, and iterative pruning results show that two same-function coordinates can select different masks and produce different post-pruning validation accuracy or perplexity before any retraining occurs, and the SwiGLU MLP audit extends the same issue to feed-forward channel evidence.

\subsection{Privacy, Optimizers, and Adapters}

The same distinction applies to privacy-facing, optimizer-state, and adapter uses of gradients. DP-SGD clips per-example gradients before adding noise \citep{abadi2016}, recent private-learning systems make clipping more structured through group-wise, per-layer, or adaptive thresholds \citep{he2023groupwise,xia2023adaptive}, and private fine-tuning of large pretrained language models, where the clipping threshold is a tuned and utility-critical hyperparameter, is the workflow our clipping audits target \citep{li2022dplm}. Adam \citep{kingma2015adam} is explicitly motivated by invariance to diagonal rescaling of the gradients, and AdamW \citep{loshchilov2019} builds coordinate-wise state from gradient histories, so that rescaling invariance is a statement about the chosen coordinates rather than about the represented function.

LoRA adds a further identifiability issue because \citet{hu2022} parameterize an update through factors whose product is used by the forward pass. Recent variants apply factor-specific learning-rate scaling or rank-allocation and pruning signals to those factors \citep{hayou2024loraplus,zhang2023adalora,liu2024alora}. Other work preconditions factor updates with a Riemannian metric \citep{zhang2024riemannianlora} or argues that the factors should be trained differently \citep{zhu2024asymmetry}. The sensitivity-based rank budgeting of \citet{zhang2023adalora} softly restricts the factor gauge through an orthogonality penalty, an approximate version of the identifiability requirement that Proposition~\ref{prop:rank-coordinate-preserving} characterizes exactly. GRR does not propose a better clipping rule, optimizer, or rank allocator. The audit instead tests whether the per-example, optimizer-state, and rank-coordinate evidence such workflows consume is identifiable, and our results show that clipping decisions, AdamW paired-product updates, and rank-coordinate importance can all change while the predictions are identical.

\subsection{Reliability Tests for Explanations}

The audit is also related to reliability tests for explanations and saliency. \citet{sundararajan2017} motivate integrated gradients with axioms such as implementation invariance, and \citet{adebayo2018} show through model-parameter randomization tests that some saliency methods fail basic sanity checks. Closest in spirit, \citet{kindermans2019} show that some saliency methods are not invariant to a constant input shift that leaves the network output unchanged, an input-side analogue of the parameter-side invariance tested here. Inside Transformers, \citet{jain2019attention} construct substantially different attention distributions that leave predictions essentially unchanged, so an internal quantity that reads as an explanation need not be pinned down by model behavior, and \citet{wiegreffe2019} qualify that conclusion by showing that the verdict depends on what an explanation is required to do, which is one reason GRR reports a concrete downstream decision instead of an interpretive claim.

The same reliability concern extends to training-data attribution, where gradient-based influence estimates are used to explain and debug individual predictions \citep{koh2017}, are scaled to large language models through curvature approximations \citep{grosse2023influence}, and have also been found fragile in deep non-convex models \citep{basu2021influence}. GRR applies the same reliability philosophy to parameter-gradient evidence, moving along an exact same-function symmetry orbit and testing whether pruning, clipping, or rank-selection conclusions are preserved.

\section{Exact Symmetry Details, Remedies, and Proofs}
\label{app:proofs}

\subsection{Grouped-Query Attention and Exactness Conditions}

The gauge-invariant objects of a head are therefore the paired products $W_{Q,h}W_{K,h}^\top$ and $W_{V,h}W_{O,h}$, which are also the combined query--key and value--output maps used in circuit-level descriptions of attention \citep{elhage2021circuits}. The individual projection matrices are only one coordinate representation of those maps, so any head-level measurement computed from a single projection factor is a statement about the representation rather than about the head.

GQA \citep{ainslie2023gqa} restricts the admissible gauges but does not remove those gauges. If $m$ query heads share one key and value head, exact preservation requires the same scalar $c_g$ for all query heads in group $g$:
\begin{equation}
    W'_{Q,h}\coloneqq c_g W_{Q,h},\quad W'_{K,g}\coloneqq c_g^{-1}W_{K,g}, \qquad h\in g,
\end{equation}
and similarly $W'_{V,g}\coloneqq c_gW_{V,g}$ with every output-projection slice for $h\in g$ scaled by $c_g^{-1}$. Our GQA audit uses this group-wise gauge.

\paragraph{Exactness conditions.}
The attention gauges above are exact in real arithmetic under the same attention mask, the same dropout-disabled evaluation graph, and the corresponding bias transformations. Scalar Q--K scaling also commutes with RoPE rotations, which is why the Q--K gauge remains valid in the Llama-style audit, while arbitrary non-scalar Q--K mixing would generally be restricted by RoPE and by GQA sharing. V--O scaling is independent of the positional mechanism because the V--O gauge rescales the value branch and the matching output-projection slice after the attention probabilities have been computed. Fused kernels and TF32 matmuls can introduce small finite-precision differences, so our drivers disable TF32 for preservation checks and report both max-logit difference and prediction and argmax flip rates before any gradient comparison. Our main evidence uses exact symmetries only. LayerNorm \citep{ba2016layernorm} and RMSNorm \citep{zhang2019rmsnorm} epsilons, affine parameters, and residual additions make normalization and residual-path near-symmetries implementation-dependent or only approximate, so we exclude those near-symmetries.

\paragraph{Exact attention preservation.}
For a Q--K scalar head gauge, $Q'_h=c_hQ_h$ and $K'_h=c_h^{-1}K_h$, hence $Q'_hK_h^{\prime\top}=Q_hK_h^\top$ and the softmax input is unchanged. For V--O, $V'_h=c_hV_h$ and $W'_{O,h}=c_h^{-1}W_{O,h}$, hence $P_hV'_hW'_{O,h}=P_hV_hW_{O,h}$ for the same attention matrix $P_h$. The Q, K, and V projection biases, when present, follow the respective branches; the output-projection bias is unchanged. In RoPE models, scalar $c_hI$ commutes with every RoPE rotation, which is why scalar Q--K scaling remains exact even though arbitrary $GL(d_h)$ mixing is restricted.

\subsection{Minimal Executable GRR Attention Audit}
\label{app:minimal-grr-code}

Listing~\ref{lst:minimal-grr} uses the same helpers as the experiment drivers. The listing compares raw and path-normalized choices of $R$ under a top-$k$ decision $D$; \texttt{largest=False} gives bottom-$k$ pruning. The preservation gate precedes both backward passes, and the transformation helpers transport the required projection biases. The shown validator defaults may be tightened for the numerical precision of a particular application.

\clearpage
\begin{lstlisting}[style=grrpython,caption={Minimal executable GRR audit for a standard attention pair. The functions imported from \texttt{grr} are the same helpers used by the full experiment drivers.},label={lst:minimal-grr}]
from collections.abc import Sequence
from typing import Literal
import torch
from torch import nn
import grr.evaluation as E
import grr.metrics as M
import grr.transforms as T
from grr.data import EncodedBatch
from grr.modeling import clone_model

def minimal_grr_attention_audit(
    base_model: nn.Module, audit_batches: Sequence[EncodedBatch],
    device: torch.device, layer_index: int,
    pair: Literal["qk", "vo"],
    scales: Sequence[float],
    *,
    k: int = 3,
    largest: bool = True,
    max_forward_delta: float = 1.0e-2,
    max_flip_rate: float = 1.0e-3,
) -> dict[str, float]:
    if len(audit_batches) == 0:
        raise ValueError("audit_batches must be nonempty")
    original = clone_model(base_model, device)
    transformed = clone_model(base_model, device)
    if pair not in {"qk", "vo"}:
        raise ValueError(f"unknown attention pair: {pair}")
    parts = {"qk": ("q", "k"), "vo": ("v", "o")}[pair]
    apply_gauge = {
        "qk": T.apply_qk_head_scaling,
        "vo": T.apply_vo_head_scaling,
    }[pair]
    apply_gauge(transformed, layer_index, scales)

    # The preservation null must pass before gradient evidence is used.
    delta_f = E.max_logit_delta(
        original, transformed, audit_batches, device)
    flip_rate = E.prediction_flip_rate(
        original, transformed, audit_batches, device)
    if delta_f > max_forward_delta or flip_rate > max_flip_rate:
        raise RuntimeError(
            "invalid GRR audit: forward preservation failed")

    E.backward_on_batches(original, audit_batches, device)
    E.backward_on_batches(transformed, audit_batches, device)

    def paired_scores(score_fn):
        return (score_fn(original, layer_index, parts),
                score_fn(transformed, layer_index, parts))

    raw_0, raw_1 = paired_scores(E.attention_head_gradient_scores)
    path_0, path_1 = paired_scores(
        E.attention_head_path_normalized_gradient_scores)

    if not 1 <= k <= len(raw_0):
        raise ValueError(f"k must lie in [1, {len(raw_0)}]")
    raw_d0 = M.topk_indices(raw_0, k, largest=largest)
    raw_d1 = M.topk_indices(raw_1, k, largest=largest)
    path_d0 = M.topk_indices(path_0, k, largest=largest)
    path_d1 = M.topk_indices(path_1, k, largest=largest)

    return {
        "forward_max_abs_delta": delta_f,
        "prediction_flip_rate": flip_rate,
        "raw_spearman": M.spearman(raw_0, raw_1),
        "raw_decision_overlap": len(raw_d0 & raw_d1) / k,
        "raw_decision_jaccard": M.jaccard(raw_d0, raw_d1),
        "path_spearman": M.spearman(path_0, path_1),
        "path_decision_overlap": len(path_d0 & path_d1) / k,
        "path_decision_jaccard": M.jaccard(path_d0, path_d1),
    }
\end{lstlisting}

\subsection{SwiGLU MLP Up--Down Symmetry}

Modern Llama-style decoder blocks \citep{touvron2023llama} also contain a non-attention bilinear symmetry in the SwiGLU MLP \citep{shazeer2020}. With row-vector notation,
\begin{equation}
    \operatorname{MLP}(x) \coloneqq (\operatorname{SiLU}(xW_g+b_g)\odot (xW_u+b_u))W_d+b_d,
\end{equation}
where $W_g$ is the gate projection, $W_u$ the up projection that supplies the value branch, and $W_d$ the down projection. Llama-style implementations typically omit these biases. In this row-vector convention, let $W_{u,j}$ denote column $j$ of $W_u$ and $W_{d,j}$ denote row $j$ of $W_d$. For each intermediate channel $j$ and scalar $c_j>0$,
\begin{equation}
    W'_{u,j}\coloneqq c_j W_{u,j}, \qquad b'_{u,j}\coloneqq c_j b_{u,j}, \qquad W'_{d,j}\coloneqq c_j^{-1}W_{d,j}
\end{equation}
preserves the channel contribution because the post-gate activation for channel $j$ is multiplied by $c_j$ and the corresponding down-projection row is divided by $c_j$. The gate branch and down-projection bias are unchanged, while only the optional up-projection bias follows the up branch. This up--down gauge gives a direct channel-level analogue of the V--O gauge and lets us test whether the GRR failure is confined to attention heads.

\paragraph{Exact SwiGLU up--down preservation.}
For an intermediate channel $j$ in a SwiGLU MLP, write $a_j(x)\coloneqq\operatorname{SiLU}(xW_{g,j}+b_{g,j})(xW_{u,j}+b_{u,j})$ and output contribution $a_j(x)W_{d,j}$, where $W_{u,j}$ is a column and $W_{d,j}$ is a row in the row-vector convention of the main text. Under $W'_{u,j}=c_jW_{u,j}$, $b'_{u,j}=c_jb_{u,j}$, and $W'_{d,j}=c_j^{-1}W_{d,j}$, the gate term is unchanged and $a'_j(x)=c_j a_j(x)$. Therefore $a'_j(x)W'_{d,j}=a_j(x)W_{d,j}$ for every input, while the unchanged down-projection bias is added afterward. The same chain-rule calculation as the bilinear case gives $\nabla_{W'_{u,j}}\cL=c_j^{-1}\nabla_{W_{u,j}}\cL$ and $\nabla_{W'_{d,j}}\cL=c_j\nabla_{W_{d,j}}\cL$, with the optional up-bias gradient following the up-weight gradient. Thus the raw up-column and down-row gradients scale oppositely. The transposed storage convention of common linear-layer implementations represents the up-column and down-row gradients as an up row and a down column, respectively. Euclidean channel-gradient norms and clipping thresholds therefore need not be invariant even though the MLP function is unchanged.

\subsection{Product-Space Displacement under Scalar Gauges}

\begin{proposition}[Exact product-space displacement under a scalar gauge]
\label{prop:product-step-identity}
Let $\cL$ be differentiable and suppose that, for every $c>0$, the reciprocal action $A'\coloneqq c^{-1}A$, $B'\coloneqq cB$ is an exact function-preserving reparameterization and leaves the audited loss invariant. Evaluate all unprimed gradients at $(B,A)$ and set $M\coloneqq BA$. Let $U\coloneqq B\nabla_A\cL$ and $V\coloneqq\nabla_B\cL A$ be the two first-order product tangents induced by a Euclidean factor-gradient step. At the same-function representative $(B',A')$, apply learning rate $\eta>0$ in the primed coordinates and define $\delta M_c\coloneqq B'_+A'_+-M$. Then
\begin{equation}
    \delta M_c = -\eta(c^2 U+c^{-2}V) +\eta^2\nabla_B\cL\nabla_A\cL.
\end{equation}
\end{proposition}

\begin{corollary}[Unbounded product-space displacement]
\label{cor:product-step-dependence}
Under the conditions and notation of Proposition~\ref{prop:product-step-identity}, if either $U$ or $V$ is nonzero, $\norm{\delta M_c}_F$ can be made arbitrarily large by changing $c$ without changing the pre-update function. Consequently, neither the exact finite product-space displacement nor the first-order term $-\eta(c^2U+c^{-2}V)$ is constant on the same-product scalar-gauge orbit, and hence neither is determined by $M$ alone.
\end{corollary}

\begin{corollary}[Limiting directions and direction non-identifiability]
\label{cor:product-step-directions}
Under the conditions and notation of Proposition~\ref{prop:product-step-identity}, if $U\neq0$, then
\begin{equation}
    \frac{\delta M_c}{\norm{\delta M_c}_F} \longrightarrow -\frac{U}{\norm{U}_F} \qquad\text{as }c\to\infty.
\end{equation}
If $V\neq0$, then
\begin{equation}
    \frac{\delta M_c}{\norm{\delta M_c}_F} \longrightarrow -\frac{V}{\norm{V}_F} \qquad\text{as }c\to0.
\end{equation}
In particular, if both are nonzero and $U/\norm{U}_F\neq V/\norm{V}_F$, the same-product orbit has two distinct limiting normalized displacement directions.
\end{corollary}

Corollary~\ref{cor:product-step-dependence} gives the magnitude failure, while Corollary~\ref{cor:product-step-directions} identifies the two asymptotic directions and gives an explicit condition under which the two directions differ. This update-level dependence is the mechanism behind our paired-product tangent measurements. In the original coordinates, with $A_+\coloneqq A-\eta\nabla_A\cL$ and $B_+\coloneqq B-\eta\nabla_B\cL$, one Euclidean factor-gradient step gives the exact product change
\begin{equation}
    B_+A_+-BA =-\eta(B\nabla_A\cL+\nabla_B\cL A) +\eta^2\nabla_B\cL\nabla_A\cL,
\end{equation}
whereas, after scalar gauge, the two first-order terms are reweighted by $c^2$ and $c^{-2}$ and the second-order term is unchanged. The function is identical before the step, but the same Euclidean gradient-descent rule in the new coordinates can represent a different first-order product-space move.

Two conventions specify the paired-product tangent in the audited architectures. For affine projection pairs, augmenting the input by a constant coordinate includes the transformed biases in the reciprocal factors. Proposition~\ref{prop:product-step-identity} then applies to the augmented product, and the reported weight-only contraction is the corresponding weight subblock. The SwiGLU diagnostic analogously contracts the up--down weight-product subblock with the gate branch held fixed. The optional up-projection bias can be included in the augmented up factor. In these experiments, paired-product tangent denotes the first-order paired weight-product contraction $B\nabla_A\cL+\nabla_B\cL A$; the tangent omits the common factor $-\eta$ and the $\eta^2$ term and is not the complete output- or logit-space tangent. The classifier-factor and LoRA diagnostics use the full effective product directly.

\subsection{Low-Rank Coordinate Identifiability}

\begin{proposition}[Characterization of rank-coordinate-preserving gauges]
\label{prop:rank-coordinate-preserving}
Let $\Delta W=BA$ have rank $r\geq2$, let $G\in GL(r)$, and let $E_i\coloneqq e_ie_i^\top$. Under $B'\coloneqq BG$ and $A'\coloneqq G^{-1}A$, define the $i$th coordinate component $C_i(G)\coloneqq BGE_iG^{-1}A$. Then $\sum_i C_i(G)=\Delta W$. Moreover, the components agree with the original components up to one relabeling, meaning $C_i(G)=C_{\pi(i)}(I)$ for every $i$ and some permutation $\pi$, if and only if $G$ is monomial. Here a monomial matrix is the product of an invertible diagonal matrix and a permutation matrix.
\end{proposition}

\begin{corollary}[Rank-coordinate ablation is not $GL(r)$-identifiable]
\label{cor:rank-coordinate-nonidentifiability}
Under the conditions of Proposition~\ref{prop:rank-coordinate-preserving}, if $G$ is non-monomial, then for every permutation $\pi$ of $\{1,\ldots,r\}$ there exists an index $i$ such that
\begin{equation}
    C_i(G)\neq C_{\pi(i)}(I), \qquad \Delta W-C_i(G)\neq \Delta W-C_{\pi(i)}(I).
\end{equation}
Thus the family of coordinate-deletion interventions is not identifiable, even up to relabeling, under the full $GL(r)$ action, although the unablated product $\Delta W$ is preserved.
\end{corollary}

The preceding characterization and corollary are qualitative: A mixing gauge changes the coordinate components. The next result quantifies the entire orbit of a single coordinate deletion and shows that the ambiguity has no finite upper bound.

\begin{theorem}[Exact orbit range of rank-coordinate component magnitudes]
\label{thm:rank-ablation-orbit}
Let $r\geq2$, let $B\in\R^{m\times r}$ have full column rank, let $A\in\R^{r\times n}$ have full row rank, and set $M\coloneqq BA$. For $G\in GL(r)$ and $i\in\{1,\ldots,r\}$, define
\begin{equation}
    C_i(G)\coloneqq BGE_iG^{-1}A, \qquad E_i\coloneqq e_ie_i^\top.
\end{equation}
Then the exact attainable range of component magnitudes is
\begin{equation}
\label{eq:rank-ablation-range}
    \{\norm{C_i(G)}_F:G\in GL(r)\} = [\sigma_r(M),\infty),
\end{equation}
where $\sigma_r(M)>0$ is the smallest nonzero singular value of $M$.
\end{theorem}

\begin{corollary}[Unbounded coordinate-deletion effect]
\label{cor:rank-ablation-deletion}
Under the conditions of Theorem~\ref{thm:rank-ablation-orbit}, let $M_{-i}(G)\coloneqq\sum_{j\neq i}C_j(G)=M-C_i(G)$ be the product after deleting coordinate $i$. Then
\begin{align}
    \{\norm{M-M_{-i}(G)}_F:G\in GL(r)\} &=\{\norm{C_i(G)}_F:G\in GL(r)\} =[\sigma_r(M),\infty),\\
    \sup_{G\in GL(r)}\norm{M_{-i}(G)}_F&=\infty.
\end{align}
Thus a fixed labeled rank coordinate can represent an arbitrarily large product-space ablation while the unablated product $M$ remains fixed.
\end{corollary}

\subsection{Additional Symmetry-Aware Remedy Details}
\label{app:remedy-details}

\paragraph{Scalar-gauge canonicalization.}
For scalar gauges, we choose a canonical representative by balancing paired norms. For $M\coloneqq BA$ with $\norm{A}_F\norm{B}_F>0$, the scalar gauge
\begin{equation}
    B'\coloneqq cB,\qquad A'\coloneqq c^{-1}A,\qquad c\coloneqq\sqrt{\frac{\norm{A}_F}{\norm{B}_F}}
\end{equation}
sets $\norm{B'}_F=\norm{A'}_F=\sqrt{\norm{A}_F\norm{B}_F}$ exactly. We refer to the procedure in \Secref{sec:remedies} as canonical balancing, which fixes the scalar gauge. The balanced representative of \Secref{sec:remedies} is constant on each positive-scalar orbit: Balancing both $(B,A)$ and $(dB,d^{-1}A)$ for any $d>0$ produces the same pair.

In the attention remedy experiments, the canonical scale is computed from the paired weight slices, while every transformed Q, K, or V bias is transported by the same branch scale; for low-rank factors, each $B$ column is balanced against the corresponding $A$ row. The checkpoint-scale audit instead reports norms of the complete branch blocks, including projection biases when present.

\paragraph{Scalar invariance of Taylor and path-normalized scores.}

The path-normalized score is a baseline related to path-normalized and product-normalized methods. The cancellation covers reciprocal scalar gauges; the cancellation does not cover non-diagonal $GL(r)$ mixing or decisions whose components the gauge fails to preserve. The coordinate-wise score $\sum_i |\theta_i\nabla_{\theta_i}\cL|$ used in Taylor pruning \citep{molchanov2017,molchanov2019} and the block score $\norm{\theta_p}\norm{\nabla_{\theta_p}\cL}$ are stable under reciprocal scalar rescaling of paired coordinates. Each parameter and the corresponding gradient scale in opposite directions, so the scale cancels within each product: If $\theta_i'=c_i\theta_i$ and $\nabla_{\theta_i'}\cL=c_i^{-1}\nabla_{\theta_i}\cL$, then $\theta_i'\nabla_{\theta_i'}\cL=\theta_i\nabla_{\theta_i}\cL$. For a uniformly rescaled block with $\theta_p'=c\theta_p$ and $\nabla_{\theta_p'}\cL=c^{-1}\nabla_{\theta_p}\cL$, the block product is likewise unchanged, preserving the path-normalized score $I_h^{\mathrm{path}}$ of \Secref{sec:remedies}. In our experiments, these two scores act as controls. This invariance guarantee is limited to scalar rescaling: Non-diagonal $GL(r)$ mixing changes rank bases, and coordinate-wise absolute or block-normalized scores need not identify function-level components.

\paragraph{Diagonal empirical-Fisher normalization and invariance.}
As a stronger baseline related to natural-gradient and Fisher-scaled methods \citep{amari1998,martens2015}, and to diagonal Fisher scores used as per-parameter importance estimates \citep{kirkpatrick2017}, we test a diagonal empirical-Fisher-normalized head score. For coordinate $i$, let
\begin{equation}
    \widehat F_i\coloneqq\frac{1}{n}\sum_{m=1}^n (\nabla_{\theta_i}\ell_m)^2.
\end{equation}
For a head $h$ with coordinates $I_h$, define
\begin{equation}
    I_h^{\mathrm{Fisher}} \coloneqq \left( \sum_{i\in I_h} \frac{(\nabla_{\theta_i}\cL)^2}{\widehat F_i+\lambda} \right)^{1/2}.
\end{equation}
Specifically, under $\theta_i'=c_i\theta_i$, the gradient and empirical-Fisher terms obey $\nabla_{\theta_i'}\cL=c_i^{-1}\nabla_{\theta_i}\cL$ and $\widehat F_i'=c_i^{-2}\widehat F_i$. Hence every defined undamped summand is invariant when $\lambda=0$ and $\widehat F_i>0$; with fixed damping, the transformed summand is $(\nabla_{\theta_i}\cL)^2/(\widehat F_i+c_i^2\lambda)$. The score is therefore much less sensitive than a raw Euclidean norm when damping is negligible and the same examples estimate $\widehat F$. Because the empirical Fisher is not the Fisher and need not capture curvature \citep{kunstner2019}, we use the empirical Fisher as a scale-normalizing baseline rather than as a curvature model.

\paragraph{Function-space scores.}
Parameter gradients are coordinate-dependent, but the output change induced by an intervention transported to remove the same functional path is coordinate-invariant. For component $j$, define
\begin{equation}
    I_j^{\mathrm{func}} \coloneqq (\E_{x\sim\cD} \norm{z_\theta(x)-z_{\theta\setminus j}(x)}_2^2)^{1/2},
\end{equation}
where $\theta\setminus j$ removes the same head output path. This ablation score is not a raw gradient score but a function-space control used to test whether a gradient-based conclusion agrees with an invariant intervention.

\paragraph{Quotient view.}
Let a symmetry group $\cG$ act by $\phi_g(\theta)$ with $f_{\phi_g(\theta)}=f_\theta$. The orbit $\mathcal{O}_\theta\coloneqq\{\phi_g(\theta):g\in\cG\}$ should be identified as one learned function. Infinitesimal symmetry directions form the vertical space
\begin{equation}
    \mathcal{V}_\theta\coloneqq\{\xi_\theta(\Omega):\Omega\in\mathfrak{g}\}, \qquad J_\theta\xi_\theta(\Omega)=0,
\end{equation}
where $J_\theta$ is the function Jacobian on the audit distribution. At regular points where the orbit space is a smooth quotient, reliable evidence should be defined on
\begin{equation}
    T_{[\theta]}(\Theta/\cG)\simeq T_\theta\Theta/\mathcal{V}_\theta.
\end{equation}
This tangent-space identification is the standard quotient-manifold construction of Riemannian optimization on matrix manifolds \citep{absil2008}. We use that construction here as a measurement space rather than as a search space.

\subsection{Full-\texorpdfstring{$GL(r)$}{GL(r)} Bilinear Remedies}
\label{sec:full-gl-remedies}

For the bilinear factors that underlie LoRA, classifier factorizations, and admissible attention-product gauges, let
\begin{equation}
\begin{split}
    \mathcal{F}_r &\coloneqq\{(B,A)\in\R^{m\times r}\times\R^{r\times n}: \operatorname{rank}(B)=\operatorname{rank}(A)=r\},\\
    \mathcal{M}_r &\coloneqq\{M\in\R^{m\times n}:\operatorname{rank}(M)=r\}, \qquad \pi(B,A)\coloneqq BA.
\end{split}
\end{equation}

\paragraph{Bilinear fibers and \texorpdfstring{$GL(r)$}{GL(r)} orbits.}
For full-rank factors, every fiber of $\pi$ is exactly one $GL(r)$ orbit. If $BA=\widetilde B\widetilde A$ for $(B,A),(\widetilde B,\widetilde A)\in\mathcal F_r$, then there is a unique $G\in GL(r)$ such that $\widetilde B=BG$ and $\widetilde A=G^{-1}A$. Indeed, the two products have the same $r$-dimensional column space, so $B$ and $\widetilde B$ are two bases related by a unique $G$. A left inverse of $B$ then gives the relation between the $A$ factors. Hence a full-gauge-invariant quantity on $\mathcal F_r$ is determined by the product $M$.

\paragraph{Full-gauge gradient envelope.}
To complement the invariant score in \Secref{sec:remedies}, we quantify exactly how much a raw Euclidean factor-gradient norm can vary over one full-gauge orbit. The resulting envelope also separates clipping thresholds that are forced by the product differential from thresholds whose decisions can be reversed by coordinates.

\begin{theorem}[Exact full-gauge envelope of Euclidean gradients]
\label{thm:gradient-orbit-envelope}
Let $(B,A)\in\mathcal{F}_r$, let $\cL(B,A)\coloneqq\overline{\cL}(BA)$ be differentiable, and write $D_A\coloneqq\nabla_A\cL\in\R^{r\times n}$ and $D_B\coloneqq\nabla_B\cL\in\R^{m\times r}$. Assume $\operatorname{rank}(D_A)=\operatorname{rank}(D_B)=r$. For $G\in GL(r)$, set
\begin{equation}
    (B_G,A_G)\coloneqq(BG,G^{-1}A), \qquad \Gamma(G)\coloneqq (\norm{\nabla_{A_G}\cL}_F^2+ \norm{\nabla_{B_G}\cL}_F^2)^{1/2}.
\end{equation}
Then the exact attainable range over the same-product orbit is
\begin{equation}
\label{eq:gradient-orbit-range}
    \{\Gamma(G):G\in GL(r)\} = \left[ \sqrt{2\norm{D_BD_A}_*},\infty \right),
\end{equation}
where $\norm{\cdot}_*$ is the nuclear norm. In particular, the lower endpoint is full-$GL(r)$-invariant.
\end{theorem}

\begin{corollary}[Coordinate identifiability of Euclidean factor clipping]
\label{cor:clipping-orbit}
Under the conditions of Theorem~\ref{thm:gradient-orbit-envelope}, for any finite clipping threshold $\tau>0$ and the convention $b_G\coloneqq\mathbf{1}\{\Gamma(G)>\tau\}$, one has
\begin{equation}
\label{eq:clipping-orbit-dichotomy}
    b_G=1\text{ for every }G \Longleftrightarrow \tau<\sqrt{2\norm{D_BD_A}_*}.
\end{equation}
If $\tau\geq\sqrt{2\norm{D_BD_A}_*}$, both $b_G=0$ and $b_G=1$ occur at same-product representatives. Thus, outside the always-clipped regime, raw Euclidean factor clipping is not factorization-identifiable at fixed $BA$.
\end{corollary}

\paragraph{Further interpretation of the product-manifold score.}

The factor-only expression in \Eqref{eq:quotient-gradient-score} is computable without forming $Z$, and Theorem~\ref{thm:quotient-gradient} identifies the expression as a Frobenius norm, hence as a nonnegative quantity.

Rank-coordinate labels remain basis-dependent under non-diagonal gauges by Corollary~\ref{cor:rank-coordinate-nonidentifiability}. Theorem~\ref{thm:quotient-gradient} and Corollary~\ref{cor:quotient-gradient} supply a positive counterpart to the non-identifiability results: Whenever the forward loss depends on a full-rank pair only through the product of that pair, whole-product or whole-head first-order evidence can be measured by the fixed-rank product-manifold gradient norm even under non-diagonal mixing. That product-manifold score is the norm of the gradient of $\overline{\cL}|_{\mathcal M_r}$, not in general the ambient norm $\norm{Z}_F$: The normal component $(I-\Pi_B)Z(I-\Pi_A)$ is invisible to factor gradients.

The same $r\times r$ Gram factors $P$ and $Q$ appear in Riemannian preconditioners that accelerate LoRA training \citep{zhang2024riemannianlora}; here the Gram factors are used to report a gauge-invariant measurement rather than to precondition an update. Separate factor regularization, including untransported weight decay, violates the product-only assumption and must be audited as part of the optimization rule rather than absorbed into these results.

\paragraph{Practical roles and costs.}
We use function-space ablation as a control for pruning and ablation claims because the ablation targets the same functional intervention. The control costs one forward pass per component or mask, provided that the intervention still denotes the same component after relabeling. Path-normalized scores are a lower-cost screen for reciprocal scalar gauges in Q--K, V--O, and up--down pairs and require only the baseline backward pass, but such scores do not address non-diagonal $GL(r)$ mixing or gauges that change component identities. Canonical balancing requires no additional gradients and is useful for debugging plots, pruning screens, and LoRA A and B scale audits, but the procedure selects a representative coordinate rather than defining a quotient metric.

Diagonal empirical-Fisher normalization targets coordinate-wise scalar scaling and is most practical when per-example gradients are already available; the quality of that normalization depends on damping, sample noise, and neglected off-diagonal structure, and the normalized score does not cover full $GL(r)$ gauges. The exact quotient-gradient norm in Corollary~\ref{cor:quotient-gradient} handles full $GL(r)$ for identifiable full-rank products at the cost of small Gram-system solves, and that norm yields fixed-rank paired-product tangent rather than rank-coordinate evidence. Paired-product tangents require one backward pass plus product contractions and audit optimizer-step consequences in identifiable products. Finally, a GRR decision audit duplicates the score and downstream decision under a verified same-function transform. The decision audit is therefore the reliability check to run before gradient evidence is treated as functional evidence.

\paragraph{Invariant quantities.}
Input gradients of a fixed scalar function stay unchanged wherever the function is differentiable, which is why input-side attributions \citep{sundararajan2017} are outside the scope of the coordinate-dependence failure. Exact natural gradients can also be reparameterization-invariant when the metric is transported correctly \citep{martens2020}. For a bilinear product, Theorem~\ref{thm:quotient-gradient} recovers the Frobenius norm of the product-manifold gradient from the factor gradients, and Corollary~\ref{cor:quotient-gradient} shows that the product-manifold gradient norm is unchanged by any invertible change of factor basis. Taylor and path-normalized scores also stay unchanged under reciprocal scalar rescaling, but this cancellation of reciprocal scale factors does not cover general $GL(r)$ basis mixing. Appendix~\ref{app:remedy-details} gives the details. Our failure claim concerns raw Euclidean measurements and diagnostics whose approximations retain coordinate dependence.

\subsection{Scalar-Gauge Results and Proofs}
\label{app:scalar-results}

The joint score of one factor pair has a simple exact range under the same scalar gauge.

\begin{lemma}[Exact scalar-gauge score range]
\label{lem:scalar-score-range}
Let $\cL$ be differentiable and suppose that, for every $c>0$, the reciprocal action $A'\coloneqq c^{-1}A$, $B'\coloneqq cB$ is an exact function-preserving reparameterization and an exact symmetry of the audited loss, so that $\cL(B',A')=\cL(B,A)$. At a fixed representative, set $a\coloneqq\norm{\nabla_A\cL}_F^2$ and $b\coloneqq\norm{\nabla_B\cL}_F^2$, and assume $ab>0$. Define
\begin{equation}
    S(c) \coloneqq(\norm{\nabla_{A'}\cL}_F^2 +\norm{\nabla_{B'}\cL}_F^2)^{1/2}.
\end{equation}
Then the exact attainable score range is
\begin{equation}
    \{S(c):c>0\} =\left[\sqrt{2\sqrt{ab}},\infty\right).
\end{equation}
\end{lemma}

The distortion of raw gradient scores can therefore change which components appear most important. When a model contains independently gaugeable components, the same-function orbit can realize every possible raw component ranking.

\paragraph{The function does not determine the raw ranking.}
Theorem~\ref{thm:arbitrary-ranking} shows that the same model can have any raw component ranking when only the factor scales change. A component ranked last can become first, and the entire ranking can be reversed, while every prediction stays the same. Every ordering is possible because the components can be rescaled independently and each component's score has no finite upper bound. The gradients remain valid derivatives in each coordinate system, but raw gradient magnitudes do not establish which components matter to the learned function.

\begin{corollary}[Arbitrary raw top-$k$ sets]
\label{cor:arbitrary-topk}
Under the conditions of Theorem~\ref{thm:arbitrary-ranking}, for every integer $k$ with $1\leq k\leq H$, every prescribed subset of $k$ components can be made the raw top-$k$ set by same-function scales, without changing any product $B_hA_h$ or the represented function.
\end{corollary}

Corollary~\ref{cor:arbitrary-topk} extends the arbitrary-ranking result to component selection: A pipeline that takes the top-$k$ components by these raw scores can return any prescribed set along the same-function orbit. The same holds for bottom-$k$ pruning, since any set of $k$ components can occupy the lowest $k$ positions. Rescaling can therefore change which components are removed while every prediction before removal stays the same. In a workflow, such a change in selection can be an unintended consequence of rescaling factors for another purpose.

\begin{figure}[t!]
\centering
\includegraphics[width=\linewidth]{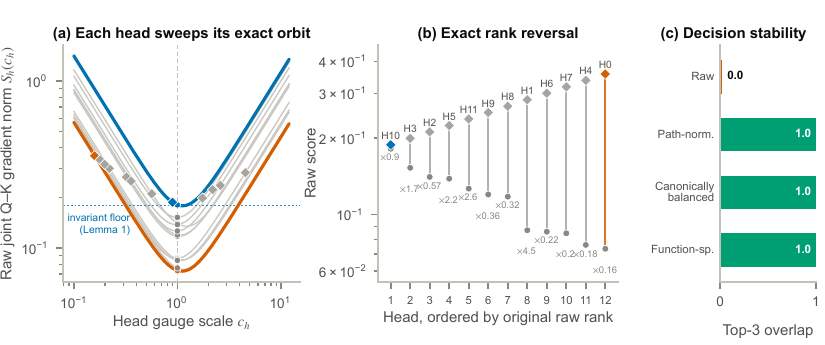}
\caption{Executing Lemma~\ref{lem:scalar-score-range} and Theorem~\ref{thm:arbitrary-ranking} on a trained checkpoint. (a) Each head's raw score follows the corresponding exact scalar-gauge orbit; circles mark the checkpoint and diamonds the constructed representative. (b) Applying the constructive proof exactly reverses the 12-head ranking. (c) Raw top-3 overlap becomes $0$, whereas path-normalized, canonically balanced, and function-space decisions are unchanged. Engineered scales lie in $[0.159,4.53]$; the maximum logit difference is $7.63\times10^{-6}$; prediction flips are $0.0\%$; and realized scores deviate from the prescribed score targets by at most $1.2\times10^{-6}$ in relative terms.}
\label{fig:qual-theory-orbit}
\end{figure}

The proof gives explicit factor scales for a prescribed raw ranking. Figure~\ref{fig:qual-theory-orbit} applies this construction of factor scales to a DistilBERT checkpoint, reversing the raw head ranking while preserving predictions and symmetry-aware conclusions.

Figure~\ref{fig:qual-theory-orbit} executes the ranking construction on a trained DistilBERT SST-2 checkpoint. The figure uses the most rank-discordant Q--K layer audit and the same 128-example audit prefix as the quantitative results.

\begin{proof}[Proof of Lemma~\ref{lem:scalar-score-range}]
The inverse coordinate change is $A=cA'$ and $B=c^{-1}B'$. The symmetry gives the function identity $\cL(B',A')=\cL(c^{-1}B',cA')$. Differentiating this loss symmetry identity with respect to $(B',A')$ and then evaluating at $(B',A')=(cB,c^{-1}A)$ gives
\begin{equation}
    \nabla_{A'}\cL=c\nabla_A\cL, \qquad \nabla_{B'}\cL=c^{-1}\nabla_B\cL.
\end{equation}
Therefore
\begin{equation}
\label{eq:ranking-score-curve}
    S(c)^2=ac^2+bc^{-2}.
\end{equation}
Set $x\coloneqq c^2>0$. The arithmetic--geometric mean inequality yields
\begin{equation}
    ax+\frac{b}{x}\geq2\sqrt{ab},
\end{equation}
with equality if and only if $x=\sqrt{b/a}$. The left-hand side is continuous on $(0,\infty)$ and tends to infinity as either $x\to0$ or $x\to\infty$. Continuity and the two limits already give the image of the left-hand side as $[2\sqrt{ab},\infty)$. For completeness, any target $t>\sqrt{2\sqrt{ab}}$ is attained explicitly by solving $ax^2-t^2x+b=0$. The discriminant $t^4-4ab$ is positive, and
\begin{equation}
    x= \frac{t^2+\sqrt{t^4-4ab}}{2a}>0
\end{equation}
gives a valid scale $c\coloneqq\sqrt{x}$. This explicit scale construction proves the exact range of $S$.
\end{proof}

\begin{proof}[Proof of Theorem~\ref{thm:arbitrary-ranking}]
For each $h$, set $a_h\coloneqq\norm{\nabla_{A_h}\cL}_F^2$ and $b_h\coloneqq\norm{\nabla_{B_h}\cL}_F^2$. Lemma~\ref{lem:scalar-score-range} applies independently to every component. Now fix any permutation $\pi$. Choose $T$ strictly larger than $\max_h\sqrt{2\sqrt{a_hb_h}}$, choose any $\varepsilon>0$, and set $t_{\pi(j)}\coloneqq T+j\varepsilon$ for $j=1,\ldots,H$. The lemma supplies an independent positive $c_{\pi(j)}$ satisfying $S_{\pi(j)}(c_{\pi(j)})=t_{\pi(j)}$. These scores have the prescribed strict ordering. Finally, $B'_hA'_h=(c_hB_h)(c_h^{-1}A_h)=B_hA_h$ for every $h$, and the assumed reciprocal actions are function-preserving. Thus all of these rankings lie on the same-function orbit.
\end{proof}

\begin{proof}[Proof of Corollary~\ref{cor:arbitrary-topk}]
Given a desired top-$k$ subset, choose a permutation whose last $k$ indices are precisely that subset and apply Theorem~\ref{thm:arbitrary-ranking}. The resulting scales preserve every product and the represented function while making that subset the raw top-$k$ set.
\end{proof}

\begin{proof}[Proof of Proposition~\ref{prop:product-step-identity}]
The loss symmetry and the same chain-rule calculation as in Lemma~\ref{lem:scalar-score-range} give $\nabla_{A'}\cL=c\nabla_A\cL$ and $\nabla_{B'}\cL=c^{-1}\nabla_B\cL$. Therefore one Euclidean gradient step in the primed coordinates is
\begin{equation}
    A'_+=c^{-1}A-\eta c\nabla_A\cL, \qquad B'_+=cB-\eta c^{-1}\nabla_B\cL.
\end{equation}
Multiplication gives the exact identity
\begin{align}
    B'_+A'_+-BA &=-\eta(c^2B\nabla_A\cL+c^{-2}\nabla_B\cL A) +\eta^2\nabla_B\cL\nabla_A\cL \\
    &=-\eta(c^2U+c^{-2}V) +\eta^2\nabla_B\cL\nabla_A\cL.
\end{align}
This expansion of the factor product gives the claimed exact displacement.
\end{proof}

\begin{proof}[Proof of Corollary~\ref{cor:product-step-dependence}]
Using Proposition~\ref{prop:product-step-identity}, set $W\coloneqq\nabla_B\cL\nabla_A\cL$. If $U\neq0$, then
\begin{equation}
    c^{-2}\delta M_c =-\eta(U+c^{-4}V)+\eta^2c^{-2}W \xrightarrow{c\to\infty} -\eta U.
\end{equation}
Taking norms in this displacement limit gives $c^{-2}\norm{\delta M_c}_F\to\eta\norm{U}_F>0$, so $\norm{\delta M_c}_F$ is unbounded. Similarly, if $V\neq0$, then
\begin{equation}
    c^2\delta M_c =-\eta(c^4U+V)+\eta^2c^2W \xrightarrow{c\to0} -\eta V,
\end{equation}
which again implies unboundedness. The same two rescalings with the $W$ term omitted show that the first-order displacement is unbounded under the same respective nonzero conditions. Since the pre-update product is $B'A'=BA$ for every $c$, neither displacement is determined by $M$ alone.
\end{proof}

\begin{proof}[Proof of Corollary~\ref{cor:product-step-directions}]
If $U\neq0$, Proposition~\ref{prop:product-step-identity} gives $c^{-2}\delta M_c\to-\eta U$ as $c\to\infty$. The limit is nonzero, so $\delta M_c\neq0$ for all sufficiently large $c$, and normalization gives
\begin{equation}
    \lim_{c\to\infty}\frac{\delta M_c}{\norm{\delta M_c}_F} =-\frac{U}{\norm{U}_F}.
\end{equation}
Similarly, if $V\neq0$, then $c^2\delta M_c\to-\eta V$ as $c\to0$, so $\delta M_c\neq0$ for all sufficiently small positive $c$ and
\begin{equation}
    \lim_{c\to0}\frac{\delta M_c}{\norm{\delta M_c}_F} =-\frac{V}{\norm{V}_F}.
\end{equation}
When both limits exist and the two normalized leading terms differ, the limiting directions are distinct, proving the final claim.
\end{proof}

\subsection{Proofs of Low-Rank Coordinate Results}

\begin{proof}[Proof of Proposition~\ref{prop:rank-coordinate-preserving}]
Because $\operatorname{rank}(BA)=r$, the matrix $B$ has full column rank and $A$ has full row rank. Hence there exist a left inverse $L_B$ of $B$ and a right inverse $R_A$ of $A$. First,
\begin{equation}
    \sum_{i=1}^r C_i(G) =BG\left(\sum_{i=1}^r E_i\right)G^{-1}A =BGG^{-1}A =BA.
\end{equation}
Suppose that the transformed components agree with the original ones up to a permutation $\pi$. Then $BGE_iG^{-1}A=BE_{\pi(i)}A$ for every $i$. Multiplying on the left by $L_B$ and on the right by $R_A$ yields
\begin{equation}
    GE_iG^{-1}=E_{\pi(i)}.
\end{equation}
The range of the matrix on the left is $G\operatorname{span}(e_i)$, whereas the range on the right is $\operatorname{span}(e_{\pi(i)})$. Therefore $Ge_i=d_i e_{\pi(i)}$ for some $d_i\neq0$. Since $\pi$ is a permutation, $G$ has exactly one nonzero entry in every row and column; that is, $G$ is monomial.

Conversely, write a monomial matrix as $G=PD$, where $P$ is a permutation matrix and $D$ is invertible diagonal. Since $D$ commutes with every $E_i$, we have $GE_iG^{-1}=PE_iP^{-1}=E_{\pi(i)}$ for the permutation represented by $P$, and consequently $C_i(G)=C_{\pi(i)}(I)$. The two directions together prove the claimed equivalence.
\end{proof}

\begin{proof}[Proof of Corollary~\ref{cor:rank-coordinate-nonidentifiability}]
For a non-monomial $G$, Proposition~\ref{prop:rank-coordinate-preserving} rules out every global relabeling that preserves all components. Hence, for each permutation $\pi$, there is an index $i$ with $C_i(G)\neq C_{\pi(i)}(I)$. Subtracting both sides from the same unablated product $\Delta W$ gives $\Delta W-C_i(G)\neq \Delta W-C_{\pi(i)}(I)$, which proves non-identifiability even up to relabeling.
\end{proof}

\begin{lemma}[Orbit of a rank-coordinate projector]
\label{lem:rank-coordinate-projector-orbit}
For any fixed $i\in\{1,\ldots,r\}$,
\begin{equation}
    \{GE_iG^{-1}\mid G\in GL(r)\} =\{uv^\top\mid u,v\in\R^r, v^\top u=1\}.
\end{equation}
\end{lemma}

\begin{proof}
For any $G\in GL(r)$, set $u\coloneqq Ge_i$ and $v\coloneqq G^{-\top}e_i$. Then $v^\top u=e_i^\top G^{-1}Ge_i=1$ and $GE_iG^{-1}=(Ge_i)(e_i^\top G^{-1})=uv^\top$. Conversely, suppose $u,v\in\R^r$ satisfy $v^\top u=1$. The vector $u$ does not belong to $\ker(v^\top)$. Choose a basis of $\ker(v^\top)$ and form a matrix $G$ whose $i$th column is $u$ and whose other columns are the chosen basis vectors, in any fixed order. These columns form a basis of $\R^r$, so $G$ is invertible. Moreover, $v^\top G=e_i^\top$, and hence $v^\top=e_i^\top G^{-1}$. Therefore $GE_iG^{-1}=uv^\top$. The two inclusions together prove the claimed set equality.
\end{proof}

\begin{proof}[Proof of Theorem~\ref{thm:rank-ablation-orbit}]
Fix $i\in\{1,\ldots,r\}$. By Lemma~\ref{lem:rank-coordinate-projector-orbit}, the remaining task is to optimize the size of $Buv^\top A$ over $u,v$ satisfying $v^\top u=1$. Define the positive-definite Gram matrices $P\coloneqq B^\top B$ and $Q\coloneqq AA^\top$. Since $Buv^\top A=(Bu)(v^\top A)$ is an outer product,
\begin{equation}
\label{eq:component-norm-gram}
    \norm{Buv^\top A}_F^2 =(u^\top Pu)(v^\top Qv).
\end{equation}
For every pair satisfying $v^\top u=1$, Cauchy--Schwarz applied to $Q^{1/2}v$ and $Q^{-1/2}u$ gives
\begin{equation}
    1=(v^\top u)^2 \leq (v^\top Qv)(u^\top Q^{-1}u).
\end{equation}
For fixed nonzero $u$, equality holds for
\begin{equation}
\label{eq:optimal-dual-vector}
    v=\frac{Q^{-1}u}{u^\top Q^{-1}u},
\end{equation}
which also satisfies $v^\top u=1$. The Gram identity in \Eqref{eq:component-norm-gram} then gives
\begin{equation}
\label{eq:component-rayleigh}
    \inf_{\substack{u,v\in\R^r\\v^\top u=1}} \norm{Buv^\top A}_F^2 = \inf_{u\neq0} \frac{u^\top Pu}{u^\top Q^{-1}u} = \lambda_{\min}(Q^{1/2}PQ^{1/2}).
\end{equation}
The last equality is the Rayleigh--Ritz characterization after the substitution $u=Q^{1/2}w$.

To identify the minimum eigenvalue, let $R_B\coloneqq BQ^{1/2}$. Then
\begin{equation}
    R_BR_B^\top=BQB^\top=BAA^\top B^\top=MM^\top, \qquad R_B^\top R_B=Q^{1/2}PQ^{1/2}.
\end{equation}
The nonzero eigenvalues of $R_BR_B^\top$ and $R_B^\top R_B$ coincide. Both have exactly $r$ positive eigenvalues because $B$ and $A$ have the stated full ranks. Hence the right-hand side of \Eqref{eq:component-rayleigh} is $\sigma_r(M)^2$. Equality is attained: Take a unit eigenvector $w_0$ for the smallest eigenvalue, set $u_0\coloneqq Q^{1/2}w_0$, and choose $v_0$ by \Eqref{eq:optimal-dual-vector}. Thus the lower endpoint in \Eqref{eq:rank-ablation-range} is attained.

We now show that every larger magnitude is attained as well. Because $r\geq2$, choose a nonzero $z\in\R^r$ with $z^\top u_0=0$, and define $v_t\coloneqq v_0+tz$ for $t\geq0$. Then $v_t^\top u_0=1$ for every $t$. Full row rank of $A$ makes $A^\top$ injective, so $A^\top z\neq0$. Full column rank of $B$ also gives $Bu_0\neq0$. Moreover, $v_0^\top Qz=(u_0^\top z)/(u_0^\top Q^{-1}u_0)=0$, so
\begin{equation}
    \norm{Bu_0v_t^\top A}_F^2 =\norm{Bu_0v_0^\top A}_F^2 +t^2\norm{Bu_0}_2^2 z^\top Qz =\sigma_r(M)^2 +t^2\norm{Bu_0}_2^2 z^\top Qz.
\end{equation}
The coefficient of $t^2$ is strictly positive, so the constructed path of rank components attains every value in $[\sigma_r(M),\infty)$. Together with the global lower bound and the orbit characterization above, this construction of rank components proves \Eqref{eq:rank-ablation-range}.
\end{proof}

\begin{proof}[Proof of Corollary~\ref{cor:rank-ablation-deletion}]
Since $\sum_jC_j(G)=M$, deleting coordinate $i$ changes the product by exactly $C_i(G)$. The first asserted range for $M-M_{-i}(G)$ follows immediately. Along any sequence for which $\norm{C_i(G)}_F\to\infty$, the reverse triangle inequality gives
\begin{equation}
    \norm{M_{-i}(G)}_F =\norm{M-C_i(G)}_F \geq \norm{C_i(G)}_F-\norm{M}_F \longrightarrow\infty,
\end{equation}
which proves the final claim.
\end{proof}

\subsection{Proofs of Full-Gauge Bilinear Results}
\label{app:full-gl-proofs}

\begin{lemma}[Exact trace range under positive-definite balancing]
\label{lem:trace-balance-range}
Let $S,T\in\R^{r\times r}$ be symmetric positive definite. Then
\begin{equation}
    \{\operatorname{tr}(SX)+\operatorname{tr}(TX^{-1}) \mid X\succ0\} = [2\operatorname{tr}((S^{1/2}TS^{1/2})^{1/2}), \infty).
\end{equation}
The minimum is attained uniquely at
\begin{equation}
    X_\star \coloneqq S^{-1/2}(S^{1/2}TS^{1/2})^{1/2}S^{-1/2}.
\end{equation}
\end{lemma}

\begin{proof}
Set $Y\coloneqq S^{1/2}XS^{1/2}$ and $R\coloneqq S^{1/2}TS^{1/2}$. Both matrices are positive definite, and
\begin{equation}
    \operatorname{tr}(SX)+\operatorname{tr}(TX^{-1}) =\operatorname{tr}(Y)+\operatorname{tr}(RY^{-1}).
\end{equation}
The following identity is an exact matrix square completion:
\begin{equation}
\label{eq:matrix-square-completion}
    \operatorname{tr}(Y)+\operatorname{tr}(RY^{-1}) -2\operatorname{tr}(R^{1/2}) = \norm{Y^{1/2}-Y^{-1/2}R^{1/2}}_F^2 \geq0.
\end{equation}
Cyclicity gives $\operatorname{tr}(R^{1/2}Y^{-1}R^{1/2})=\operatorname{tr}(RY^{-1})$, and each cross term in the squared norm has trace $\operatorname{tr}(R^{1/2})$. Equality in \Eqref{eq:matrix-square-completion} holds if and only if $Y^{1/2}=Y^{-1/2}R^{1/2}$, which, after left multiplication by $Y^{1/2}$, is equivalent to $Y=R^{1/2}$. This equality condition gives the stated unique minimizer $X_\star$.

Let $\mu\coloneqq\operatorname{tr}(R^{1/2})$. At $X_\star$, each trace term equals $\mu$. Therefore, along the ray $X=tX_\star$ with $t>0$,
\begin{equation}
    \operatorname{tr}(SX)+\operatorname{tr}(TX^{-1}) =\mu(t+t^{-1}).
\end{equation}
As $t$ ranges over $(0,\infty)$, $t+t^{-1}$ has exact range $[2,\infty)$. Together with the global lower bound, this ray construction proves the claimed range.
\end{proof}

\begin{proof}[Proof of Theorem~\ref{thm:gradient-orbit-envelope}]
At the same-product representative $(B_G,A_G)=(BG,G^{-1}A)$, the ordinary factor gradients transform as
\begin{equation}
    D_A(G)\coloneqq\nabla_{A_G}\cL=G^\top D_A, \qquad D_B(G)\coloneqq\nabla_{B_G}\cL=D_BG^{-\top}.
\end{equation}
The transformation rule follows either from the chain rule or directly by writing $D_A=B^\top Z$ and $D_B=ZA^\top$ for $Z\coloneqq\nabla_M\overline{\cL}(BA)$. In particular,
\begin{equation}
    D_B(G)D_A(G)=D_BG^{-\top}G^\top D_A=D_BD_A=ZM^\top Z,
\end{equation}
so the matrix appearing in the lower endpoint of \Eqref{eq:gradient-orbit-range} is gauge-invariant.

The full-rank assumptions make $S\coloneqq D_AD_A^\top$ and $T\coloneqq D_B^\top D_B$ positive definite. Let $X\coloneqq GG^\top$. A cyclic-trace calculation gives
\begin{equation}
\label{eq:gradient-orbit-objective}
    \Gamma(G)^2 =\norm{G^\top D_A}_F^2 +\norm{D_BG^{-\top}}_F^2 =\operatorname{tr}(SX)+\operatorname{tr}(TX^{-1}).
\end{equation}
Conversely, every positive-definite $X$ has a factorization $X=GG^\top$ with $G\in GL(r)$. Thus ranging over gauges is equivalent to ranging the right-hand side of \Eqref{eq:gradient-orbit-objective} over $X\succ0$.

By Lemma~\ref{lem:trace-balance-range}, the squared joint gradient norm has exact range
\begin{equation}
    [2\operatorname{tr}(R^{1/2}),\infty), \qquad R\coloneqq S^{1/2}TS^{1/2}.
\end{equation}
The remaining step is to identify the lower endpoint. The squared singular values of $D_BD_A$ are the positive eigenvalues of $(D_BS^{1/2})(D_BS^{1/2})^\top$, which coincide with the eigenvalues of $(D_BS^{1/2})^\top(D_BS^{1/2})=S^{1/2}TS^{1/2}=R$. Hence
\begin{equation}
    \norm{D_BD_A}_*=\operatorname{tr}(R^{1/2}),
\end{equation}
so taking square roots of the range supplied by the lemma gives exactly \Eqref{eq:gradient-orbit-range}.
\end{proof}

\begin{proof}[Proof of Corollary~\ref{cor:clipping-orbit}]
For a finite threshold $\tau$, if $\tau<\sqrt{2\norm{D_BD_A}_*}$, the lower bound forces every representative to be clipped. If the reverse weak inequality holds, a minimizing gauge has $\Gamma(G)\leq\tau$ and is not clipped, while the unbounded part of \Eqref{eq:gradient-orbit-range} supplies another gauge with $\Gamma(G)>\tau$. These two cases prove \Eqref{eq:clipping-orbit-dichotomy} and the stated two-sided attainability.
\end{proof}

\begin{proof}[Proof of Theorem~\ref{thm:quotient-gradient}]
Let $Z\coloneqq\nabla_M\overline{\cL}(M)$. The product differential and the Frobenius inner product give
\begin{equation}
    d\cL[B,A][dB,dA] =\langle Z,(dB)A+B(dA)\rangle_F =\langle ZA^\top,dB\rangle_F +\langle B^\top Z,dA\rangle_F.
\end{equation}
Consequently,
\begin{equation}
    D_A=B^\top Z, \qquad D_B=ZA^\top, \qquad D_AA^\top=B^\top D_B=B^\top ZA^\top=\Psi.
\end{equation}

For completeness, the differential of the product map has image
\begin{equation}
    T_M\mathcal{M}_r =\{BX+YA\mid X\in\R^{r\times n}, Y\in\R^{m\times r}\}.
\end{equation}
The kernel of that differential is $\{(B\Omega,-\Omega A):\Omega\in\R^{r\times r}\}$, so the image has dimension $mr+rn-r^2=r(m+n-r)=\dim\mathcal{M}_r$ and is the full tangent space. The matrices $\Pi_B$ and $\Pi_A$ are the orthogonal projectors onto $\operatorname{col}(B)$ and $\operatorname{row}(A)$. A matrix $W$ is Frobenius-orthogonal to every tangent if and only if $B^\top W=0$ and $WA^\top=0$. Hence the normal projection of $Z$ is $(I-\Pi_B)Z(I-\Pi_A)$, and the tangent projection is
\begin{equation}
    \Pi_{T_M\mathcal{M}_r}Z =\Pi_BZ+Z\Pi_A-\Pi_BZ\Pi_A.
\end{equation}
For every tangent $\dot M$, orthogonality of the projection gives
\begin{equation}
    d\overline{\cL}(M)[\dot M] =\langle Z,\dot M\rangle_F =\langle\Pi_{T_M\mathcal{M}_r}Z,\dot M\rangle_F.
\end{equation}
Thus this tangent projection is exactly the Riemannian gradient of the restriction of $\overline{\cL}$ to $\mathcal{M}_r$ under the Frobenius metric.

The three blocks $\Pi_BZ\Pi_A$, $\Pi_BZ(I-\Pi_A)$, and $(I-\Pi_B)Z\Pi_A$ are pairwise orthogonal in the Frobenius inner product. This pairwise Frobenius orthogonality gives
\begin{equation}
\label{eq:tangent-norm-decomposition}
    \norm{\Pi_{T_M\mathcal{M}_r}Z}_F^2 =\norm{\Pi_BZ}_F^2+\norm{Z\Pi_A}_F^2 -\norm{\Pi_BZ\Pi_A}_F^2.
\end{equation}
Using idempotence and symmetry of the projectors together with the factor-gradient identities above, the first two terms are
\begin{align}
    \norm{\Pi_BZ}_F^2 &=\operatorname{tr}(P^{-1}B^\top ZZ^\top B) =\operatorname{tr}(P^{-1}D_AD_A^\top),\\
    \norm{Z\Pi_A}_F^2 &=\operatorname{tr}(Q^{-1}AZ^\top ZA^\top) =\operatorname{tr}(Q^{-1}D_B^\top D_B).
\end{align}
Moreover, $\Pi_BZ\Pi_A=BP^{-1}\Psi Q^{-1}A$, so a direct cyclic-trace calculation yields
\begin{equation}
    \norm{\Pi_BZ\Pi_A}_F^2 =\operatorname{tr}(Q^{-1}\Psi^\top P^{-1}\Psi).
\end{equation}
Substitution into \Eqref{eq:tangent-norm-decomposition} proves both \Eqref{eq:quotient-gradient-score} and \Eqref{eq:quotient-gradient-projection}.
\end{proof}

\begin{proof}[Proof of Corollary~\ref{cor:quotient-gradient}]
Under $(B,A)\mapsto(BG,G^{-1}A)$, the product $M$ and hence $Z$ are unchanged. In addition, direct substitution shows $\Pi_{BG}=\Pi_B$ and $\Pi_{G^{-1}A}=\Pi_A$. The projected-gradient norm is therefore invariant under every $G\in GL(r)$, and the factor-only expression is equal to that same invariant norm. Finally, the norm vanishes if and only if $Z$ lies in the normal space, equivalently if and only if $B^\top Z=0$ and $ZA^\top=0$. By the factor-gradient identities in the proof of Theorem~\ref{thm:quotient-gradient}, these conditions are exactly $D_A=D_B=0$.
\end{proof}

\section{Additional Experimental Details}
\label{app:additional-experiments}

\subsection{Setup}

We evaluate on SST-2, MNLI, QNLI, and MRPC from GLUE \citep{wang2018}. SST-2 is based on the Stanford Sentiment Treebank \citep{socher2013}, MNLI on the Multi-Genre Natural Language Inference corpus \citep{williams2018}, and MRPC on the Microsoft Research Paraphrase Corpus \citep{dolan2005}. We use public fine-tuned checkpoints and six independently trained LoRA adapters. For SST-2, we audit all attention layers of DistilBERT \citep{sanh2019} and BERT-base \citep{devlin2019}, together with selected early, middle, and late layers of ELECTRA-base \citep{clark2020} and RoBERTa-base \citep{liu2019}, in the main GLUE sweep. For MNLI, QNLI, and MRPC, we audit selected early, middle, and late RoBERTa-base layers. MNLI labels are remapped to the RoBERTa convention of contradiction, neutral, and entailment.

GLUE validation accuracy and pruning accuracy use full validation splits. Accuracy differences between decisions made in two same-function coordinates are reported in percentage points, abbreviated pp in tables. Gradient-score audits use 128 validation examples, and per-example clipping audits use 48 examples. To test statistical robustness, we run 324 DistilBERT SST-2 audits over all six layers, shuffled validation subsets of size 32, 128, and 256, three shuffled seeds, Q--K and V--O pairs, and mild, medium, and aggressive gauge ranges. We run all-layer 96-audit checks for BERT-base, ELECTRA-base, and RoBERTa on SST-2, as well as RoBERTa on MNLI, QNLI, and MRPC, with subset size 128, two shuffled seeds, Q--K and V--O pairs, and mild and aggressive gauge ranges.

Because clipping decisions are especially sensitive to audit-set size, a separate 216-audit DistilBERT on SST-2 clipping robustness suite covers all six layers, 48- and 128-example shuffled subsets, three shuffled seeds, Q--K and V--O pairs, and mild, medium, and aggressive gauge ranges. A 256-example clipping holdout covers early, middle, and late DistilBERT layers, a held-out seed, both attention pairs, and all three gauge ranges. Appendix~\ref{app:repro} records the exact seed value used by each suite.

As a cross-domain check, we also audit all 12 layers of a ViT-base checkpoint fine-tuned on CIFAR-10 \citep{dosovitskiy2021,krizhevsky2009}, using the Hugging Face identifier \path{nateraw/vit-base-patch16-224-cifar10}. We evaluate accuracy and pruning effects on the full 10,000-image CIFAR-10 test split, compute gradients on 256 test images, and compute per-example clipping on 48 images. The ViT experiment is reported separately from the GLUE aggregate because the modality differs.

To test whether the phenomenon extends beyond encoder classification, we audit decoder-only next-token language modeling on WikiText-2 \citep{merity2017}. We use GPT-2-medium and GPT-2-large \citep{radford2019} and a Llama-style \citep{touvron2023llama} GQA model, \path{TinyLlama/TinyLlama-1.1B-intermediate-step-1431k-3T} \citep{zhang2024tinyllama}. For GPT-2-medium, we audit all 24 layers, evaluate perplexity on 128 contiguous 128-token test blocks, compute gradients on 32 validation blocks, and compute per-sequence clipping on 16 validation blocks. A GPT-2-medium global V--O pruning profile covers all 384 heads with mild, medium, and aggressive gauges, a 15\% global head budget, the same 128 test blocks, and 32 validation audit blocks.

The remaining decoder-only audits follow the same structure at smaller evaluation budgets. For GPT-2-large, we audit all 36 layers with 64 test blocks, 16 validation audit blocks, and eight per-sequence clipping blocks, together with a global V--O pruning profile over all 720 heads under the same profile family and a 15\% budget. For TinyLlama, we audit all 22 attention layers with 64 test blocks, eight validation audit blocks, and eight per-sequence clipping blocks; the 32 TinyLlama query heads share four key and value heads, so the audit uses the group-wise gauge described above. We also audit the TinyLlama SwiGLU MLP up--down symmetry on layers 0, 11, and 21 using the same WikiText-2 blocks and top-128 channel rankings over all 5,632 intermediate channels. Because exact forward preservation is the audit precondition, the experiment drivers disable TF32 matrix multiplication and use FP32 matmuls for the preservation checks.

The experiment code is organized as a single audit suite. Separate drivers run the main GLUE audit, the trained LoRA audits and rank-pruning recovery curves, the subset and scale robustness suite, the clipping robustness suite, the ViT layer-level audit, the ViT full-test global-pruning calibration, the decoder-only WikiText-2 attention, global-pruning, and MLP audits, the one-shot and iterative pruning pipelines, profile-wise global-pruning scale calibrations for RoBERTa on SST-2, QNLI, and MRPC, the downstream-remedy audit, the AdamW optimizer-state audit, the checkpoint-scale audit, and the training-time imbalance calibration.

A multi-GPU suite runner can execute these drivers as independent GPU-isolated shards and merge the shard outputs into a single result table, while a validation script checks the reported invariants: baseline quality, forward-preservation tolerances for logit drift and prediction and argmax flip rates, finite numeric fields, unique result keys, and expected evidence rows including the ViT and cross-task global-pruning profiles. A lightweight summarization script reconstructs the reported aggregates directly from that result table. The runs were executed on a 4$\times$H100 node; each audit uses deterministic seeds and records the exact model tag, task, transform, metric, and sample count in the table.

\begingroup\small\raggedright The exact public model identifiers are \path{distilbert-base-uncased-finetuned-sst-2-english}, \path{textattack/bert-base-uncased-SST-2}, \path{howey/electra-base-sst2}, \path{textattack/roberta-base-SST-2}, \path{textattack/roberta-base-MNLI}, \path{textattack/roberta-base-QNLI}, \path{textattack/roberta-base-MRPC}, \path{nateraw/vit-base-patch16-224-cifar10}, \path{gpt2-medium}, \path{gpt2-large}, and \path{TinyLlama/TinyLlama-1.1B-intermediate-step-1431k-3T}. The trained LoRA experiments start from \path{distilbert-base-uncased} for SST-2 and \path{textattack/roberta-base-MRPC} for MRPC. GLUE splits, MNLI label remapping, audit subset sizes, gauge scales, pruning fractions, and fine-tuning recipes are encoded in the experiment drivers and serialized into a companion run-details record. Every table is therefore traceable to a row filter over the result table.
\par\endgroup

\begin{table}[t!]
\centering
\caption{Real-task baselines and audit scale. Here and in every later table, GLUE accuracies use full validation splits, CIFAR-10 uses the full 10,000-image test split, and WikiText-2 uses fixed public evaluation subsets for the decoder-only audits. Global-profile sweeps are reported separately and are not included in the audit counts below.}
\label{tab:baseline}
\small
\resizebox{\linewidth}{!}{%
\begin{tabular}{llccc}
\toprule
Model & Task & Metric & Audit count & Eval $n$ \\
\midrule
DistilBERT fine-tuned & \multirow{4}{*}{SST-2} & 91.06\% & 12 & \multirow{4}{*}{872} \\
BERT-base fine-tuned & & 92.43\% & 24 & \\
RoBERTa-base fine-tuned & & 94.04\% & 6 & \\
ELECTRA-base fine-tuned & & 95.18\% & 6 & \\
\addlinespace RoBERTa-base fine-tuned & MNLI-m & 88.10\% & 6 & 9,815 \\
RoBERTa-base fine-tuned & QNLI & 92.68\% & 6 & 5,463 \\
RoBERTa-base fine-tuned & MRPC & 91.18\% & 6 & 408 \\
\addlinespace DistilBERT rank-16 LoRA, 3 seeds & SST-2 & $89.87\%$ $(89.68\%,90.25\%)$ & 3 adapter audits & 872 each \\
RoBERTa rank-8 LoRA, 3 seeds & MRPC & $90.28\%$ $(89.95\%,90.69\%)$ & 3 adapter audits & 408 each \\
\midrule
ViT-base fine-tuned & CIFAR-10 & 98.51\% & 24 & 10,000 \\
\midrule
GPT-2-medium pretrained & \multirow{4}{*}{WikiText-2} & ppl. 46.75 & 48 & 16,256 tokens \\
GPT-2-large pretrained & & ppl. 45.07 & 72 & \multirow{3}{*}{8,128 tokens} \\
TinyLlama 1.1B GQA pretrained & & ppl. 18.77 & 44 & \\
TinyLlama 1.1B SwiGLU MLP & & ppl. 18.77 & 3 MLP audits & \\
\midrule
Total direct attention-pair audits & All tasks & \multirow{2}{*}{Count} & 254 & Varies \\
Additional MLP channel audits & WikiText-2 & & 3 & 8,128 tokens \\
\bottomrule
\end{tabular}
}
\end{table}

\subsection{Audit Roadmap and Workflow Configuration}

Every experiment begins from numerically verified same-function coordinates. The layer-level audits test score, ranking, clipping, and downstream-decision stability; the global and iterative pruning and LoRA studies test complete workflows. The ViT, decoder-only, GQA, and SwiGLU audits extend the same protocol across modalities and architectures, while checkpoint-scale and training-time measurements calibrate the stress-test gauges against naturally occurring imbalance. Function-space ablation, path normalization, canonical balancing, diagonal Fisher normalization, and the ViT input-pixel gradient serve as controls for the symmetries or coordinate systems each control targets. The analytical range and quotient results in Appendix~\ref{app:proofs} characterize the corresponding worst cases and invariant alternatives.

\begin{figure}[t!]
\centering
\includegraphics[width=\linewidth]{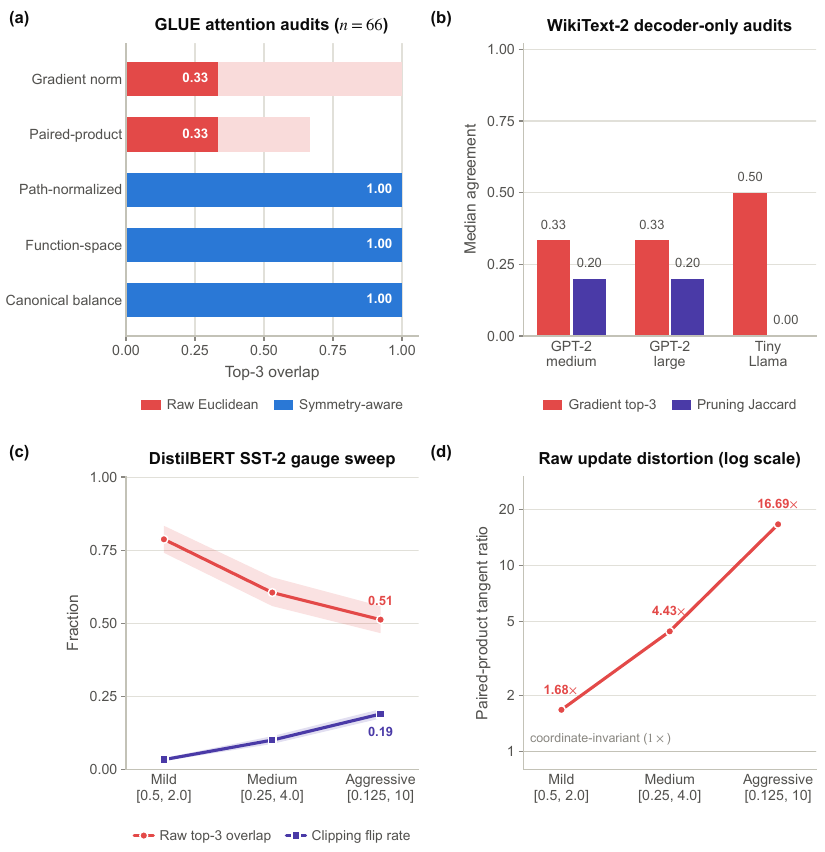}
\caption{Summary GRR diagnostics. (a) Raw GLUE attention gradients and paired-product tangents have low same-function top-3 agreement, whereas path-normalized, function-space, and canonically balanced scores are stable under the scalar gauges tested; bars show medians and pale bands show observed ranges. (b) Decoder-only raw rankings and pruning masks also have low agreement. The scale sweeps in (c) and (d) show that DistilBERT stability decreases smoothly as the gauge range grows, with zero prediction flips; shading gives bootstrap 95\% confidence intervals for the mean.}
\label{fig:summary}
\end{figure}

Figure~\ref{fig:summary} summarizes the cross-setting empirical pattern and the scalar-gauge controls used throughout these audits.

The main 12-head GLUE and ViT layer audits and one-shot global audits use the aggressive scale vector
\begin{equation}
(0.125,0.20,0.35,0.50,0.75,1.25,1.75,2.50,3.50,5.00,7.50,10.00).
\end{equation}
The robustness sweeps and experiments with other numbers of heads, shared groups, or channels use geometric spacing over the applicable mild $[0.5,2.0]$, medium $[0.25,4.0]$, or aggressive $[0.125,10.0]$ range. The 12-head global-profile driver uses the same 12-head scale vector for the aggressive case. For classifier factors, we use both diagonal gauges and non-diagonal $GL(r)$ gauges.

For the trained LoRA adapters, we train DistilBERT-base on SST-2 for 3 epochs under three training seeds with rank 16, $\alpha=32$, dropout 0.05, AdamW, linear warmup, and LoRA modules on the Q, K, V, and O projections. We also train rank-8 RoBERTa adapters on MRPC from the public \path{textattack/roberta-base-MRPC} checkpoint with $\alpha=16$, the same three seeds, LoRA modules on the Q, K, and V projections, and the classifier saved with the adapter.

As a more pipeline-like pruning check, we run four rounds of iterative global V--O pruning or top-head ablation with RoBERTa on MRPC. Each round recomputes raw gradients on the current masked model, selects seven additional heads out of 144, zeros the output paths of those heads, and evaluates on the full validation split. The original, reparameterized, and canonically balanced runs start from same-function coordinates and use the same audit subset.

For LoRA rank-pruning recovery, each seed uses the bottom-25\% rank coordinates selected by the original and reparameterized raw rank-gradient scores. Both selected rank sets are applied to the same original trained adapter, pruned LoRA ranks are kept zero by masking A rows and B columns, and the remaining trainable adapter and classifier parameters are recovered for 2 epochs. DistilBERT on SST-2 uses the full SST-2 train split with AdamW, learning rate $10^{-4}$, warmup ratio 0.06, weight decay 0.01, batch size 64, and gradient clipping at 1.0. RoBERTa on MRPC uses the full MRPC train split with learning rate $5\times 10^{-5}$ and batch size 32, matching the smaller task.

\begin{sloppypar}
Finally, profile-wise global V--O pruning calibrations cover RoBERTa on SST-2, QNLI, and MRPC, ViT on CIFAR-10, and GPT-2-medium and GPT-2-large on WikiText-2. The GLUE global-pruning calibrations use the full validation split for each task, the ViT global calibration uses the full 10,000-image CIFAR-10 test split, and the GPT-2 global calibrations use 128 contiguous WikiText-2 test blocks for GPT-2-medium and 64 blocks for GPT-2-large. All use mild, medium, and aggressive same-function V--O gauge profiles and fixed audit subsets for gradients.

The remaining audits are a ViT input-pixel gradient control, a targeted downstream-remedy audit on the layer-11 V--O pair of RoBERTa on MRPC, an AdamW optimizer-state audit, a checkpoint-scale audit, and a training-time imbalance calibration. The ViT input-gradient control applies the same three global V--O profiles and compares supervised-loss gradients with respect to normalized input pixels on 32 CIFAR-10 test images, after freezing parameter-gradient storage. The downstream-remedy audit estimates diagonal empirical-Fisher scores from 64 per-example validation gradients and computes a clipped paired-product tangent in the style of differentially private stochastic gradient descent (DP-SGD), using the original median target-pair norm as clipping threshold. The AdamW optimizer-state audit accumulates first and second moments over 512 MRPC training examples without changing parameters, compares directly accumulated transformed moments to moments transported from the original coordinate, and then compares the one-step AdamW paired-product tangent.

The checkpoint-scale audit measures the natural per-head Q--K and V--O paired norm imbalance in the eight public encoder and ViT checkpoints used above. The training-time imbalance calibration fine-tunes \path{distilbert-base-uncased} on the full SST-2 train split for 3 epochs with AdamW and records all-layer paired norm imbalance after each epoch. These calibrations quantify the mild same-function scale variation already present in trained checkpoints and in a standard full-model fine-tuning run.
\end{sloppypar}

\subsection{Forward Preservation}

Table~\ref{tab:forward} summarizes the preservation checks by transformation family. All classification, robustness, factor, and global-profile audits have zero prediction flips. Decoder-only audits also have zero next-token argmax flips except for one finite-precision GPT-2-large layer-9 Q--K case, where one audited position flips at a rate of $0.0492\%$; the median remains zero. The largest logit difference is $6.45\times10^{-3}$ for GPT-2-medium, whose audited argmax predictions are unchanged. Thus every subsequent comparison begins from a transformation that passes the forward-preservation gate, with the sole numerical exception reported explicitly.

\begin{table}[t!]
\centering
\caption{Forward preservation summary. Nonzero logit differences are finite-precision effects from rescaling trained floating-point weights. Every audit in the later tables starts from coordinates verified here, so prediction and argmax flip rates are zero unless stated otherwise.}
\label{tab:forward}
\small
\resizebox{\linewidth}{!}{%
\begin{tabular}{lccc}
\toprule
Transformation family & Audits & Max logit difference & Prediction and argmax flips \\
\midrule
GLUE layer Q--K and V--O attention gauges & 66 & $1.53\times10^{-5}$ & 0 \\
ViT attention gauges on CIFAR-10 & 24 & $1.14\times10^{-5}$ & 0 \\
ViT global V--O profiles on CIFAR-10 & 3 & $1.43\times10^{-5}$ & 0 \\
\addlinespace GPT-2-medium attention gauges on WikiText-2 & 48 & $6.45\times10^{-3}$ & 0 \\
GPT-2-large attention gauges on WikiText-2 & 72 & $9.54\times10^{-5}$ & max $0.0492\%$ \\
GPT-2-medium global V--O profiles on WikiText-2 & 3 & $3.97\times10^{-3}$ & 0 \\
GPT-2-large global V--O profiles on WikiText-2 & 3 & $1.18\times10^{-4}$ & 0 \\
TinyLlama group-wise attention gauges on WikiText-2 & 44 & $3.72\times10^{-5}$ & 0 \\
TinyLlama SwiGLU MLP gauges on WikiText-2 & 3 & $6.20\times10^{-5}$ & 0 \\
\midrule
Subset and scale robustness gauges & 900 & $2.28\times10^{-5}$ & 0 \\
Clipping robustness gauges, main and holdout & 234 & $8.58\times10^{-6}$ & 0 \\
Global all-layer gauges, including 3 ViT profiles & 26 & $1.82\times10^{-5}$ & 0 \\
\midrule
Classifier-factor gauges & 14 & $9.54\times10^{-7}$ & 0 \\
Trained LoRA factor gauges & 6 & $2.13\times10^{-5}$ & 0 \\
\bottomrule
\end{tabular}
}
\end{table}

\subsection{Low-Rank Factor Coordinates}
\label{sec:low-rank-factor-audits}

Across seven classifier heads, exact diagonal factor gauges preserve logits but change factor-coordinate evidence. Rank-gradient mean ratios have median $3.25$ and range from $3.07$ to $4.55$. Raw paired-product tangent ratios have median $14.73$ and range from $12.54$ to $21.74$. Clipping flip rates have median $39.6\%$ and range from $22.9\%$ to $50.0\%$. Function-space rank ablation under diagonal gauges is numerically invariant, with mean ratios from $0.99999994$ to $1.00000003$.

Non-diagonal $GL(r)$ gauges give another failure mode. Such gauges preserve the classifier weight but rotate the rank basis, so neither a labeled rank-coordinate component nor the corresponding coordinate-deletion intervention is identifiable. Raw rank-gradient mean ratios have median $2.76$ and range from $1.83$ to $4.11$. Paired-product tangent ratios have median $11.21$ and range from $6.69$ to $17.32$. The median clipping flip rate is $29.2\%$. Coordinate-ablation mean ratios range from $0.906$ to $1.532$. Canonical balancing returns the median gradient ratio to $1.00$ for the diagonal scalar gauge, but balancing does not resolve this full-basis ambiguity. These results show that LoRA rank importance is basis-dependent unless defined through the product or a fixed gauge.

\subsection{Trained LoRA Adapters}

We evaluate six trained LoRA adapters in two settings that vary the dataset, base model, and rank budget. Three rank-16 DistilBERT adapters are trained on SST-2 for three epochs with three seeds, yielding mean validation accuracy $89.87\%$ across runs that range from $89.68\%$ to $90.25\%$. Three rank-8 RoBERTa adapters are trained on MRPC from \path{textattack/roberta-base-MRPC} for three epochs, yielding mean accuracy $90.28\%$ across runs that range from $89.95\%$ to $90.69\%$. We audit all LoRA A and B factors in the attention projections: 384 rank coordinates per DistilBERT seed and 288 per RoBERTa seed. Exact diagonal LoRA gauges produce no prediction flips; maximum logit differences are $5.25\times10^{-6}$ and $2.13\times10^{-5}$, respectively.

The rank-coordinate evidence is consistently unstable in both settings. For DistilBERT on SST-2, raw rank-gradient top-12 overlap has median $0.417$ and paired-product tangent top-12 overlap has median $0.250$, with median tangent ratio $5.25$. For rank-8 RoBERTa on MRPC, the corresponding medians are $0.333$, $0.250$, and $5.08$. Per-example clipping decisions flip for a median $8.33\%$ of SST-2 examples and $14.58\%$ of MRPC examples. Canonicalizing A and B in both same-function coordinates restores top-12 overlap to $1.000$ for both gradient and paired-product tangent scores in all six adapters. The same dependence changes adapter-compression masks. Section~\ref{sec:lora-compression} and Table~\ref{tab:lora} report the rank-pruning, ablation, and recovery results.

\subsection{Trained LoRA Adapter Compression and Recovery}
\label{sec:lora-compression}

The coordinate dependence also changes adapter-compression decisions. Pruning the bottom 25\% of LoRA rank coordinates by raw rank-gradient score gives median rank-set Jaccard $0.401$ for DistilBERT on SST-2 and $0.455$ for RoBERTa on MRPC. Top-25\% ablation gives median Jaccard $0.297$ and $0.500$, respectively. Accuracy gaps are smaller than the rank-set disagreement: Median bottom-pruning gaps are $0.34$ percentage points for SST-2 and $0.00$ for MRPC, while median top-ablation gaps are $1.03$ and $0.25$ percentage points. The rank identities selected by raw gradient evidence are therefore not functionally identifiable, even when validation accuracy is partially buffered.

We then run rank-pruning recovery curves for the bottom-25\% masks. Both masks are applied to the same original trained adapter, so the two runs differ only in the rank sets selected by the original and reparameterized gradient scores. After two recovery epochs, the median coordinate-induced gap is $0.23$ percentage points for SST-2 and $0.00$ for MRPC; the maximum MRPC gap is $0.49$ percentage points. Recovery can absorb part of the utility loss, but recovery does not make the selected rank sets agree. Thus PEFT compression analyses based on raw factor-gradient rank coordinates should be audited or expressed through product-space quantities.

\begin{table}[h!]
\centering
\caption{Trained LoRA adapter audits over six training runs: three rank-16 DistilBERT adapters on SST-2 and three rank-8 RoBERTa adapters on MRPC.}
\label{tab:lora}
\small
\begin{tabular}{lcc}
\toprule
Measurement & DistilBERT on SST-2 & RoBERTa on MRPC \\
\midrule
Baseline accuracy & mean 89.87\% & mean 90.28\% \\
\midrule
Raw rank-gradient top-12 overlap & med. 0.417 & med. 0.333 \\
Raw paired-product tangent top-12 overlap & med. 0.250 & med. 0.250 \\
Raw paired-product tangent mean ratio & med. 5.25 & med. 5.08 \\
Canonical gradient and product top-12 overlap & 1.000, 1.000 & 1.000, 1.000 \\
Per-example clipping flip rate & med. 8.33\% & med. 14.58\% \\
\midrule
Bottom-25\% rank pruning, rank-set Jaccard & med. 0.401 & med. 0.455 \\
Bottom-25\% rank pruning, accuracy gap & med. 0.34 pp & med. 0.00 pp \\
Top-25\% rank ablation, rank-set Jaccard & med. 0.297 & med. 0.500 \\
Top-25\% rank ablation, accuracy gap & med. 1.03 pp & med. 0.25 pp \\
Bottom-25\% recovery at epoch 2, accuracy gap & med. 0.23 pp & med. 0.00 pp \\
\bottomrule
\end{tabular}
\end{table}

\subsection{Robustness to Subsets and Gauge Strength}

Table~\ref{tab:robustness} reports the multi-seed DistilBERT robustness suite over all six layers. Even mild gauges, with scales only between $0.5$ and $2.0$, reduce raw top-3 overlap on shuffled subsets to a mean of $0.787$. Medium gauges reduce the mean to $0.605$, and aggressive gauges reduce the mean to $0.512$ with bootstrap 95\% confidence interval $[0.466,0.559]$. Paired-product tangent distortion increases monotonically from $1.68\times$ to $16.69\times$. All 324 runs have zero prediction flips. Thus the failure is not an artifact of one fixed audit subset; the degradation appears across subset sizes, seeds, layers, and both attention factor pairs, and grows smoothly with gauge strength, as Figure~\ref{fig:summary}(c) and Figure~\ref{fig:summary}(d) show.

\begin{sloppypar}
The same pattern appears in six all-layer 96-audit checks on BERT-base and ELECTRA-base on SST-2, together with RoBERTa on SST-2, MNLI, QNLI, and MRPC. Table~\ref{tab:robustness_breadth} summarizes these runs. Prediction flips are again zero. Under aggressive gauges, mean raw top-3 overlaps range from $0.361$ to $0.444$ across the RoBERTa tasks, are $0.424$ for BERT on SST-2, and are $0.444$ for ELECTRA on SST-2, while mean paired-product tangent ratios remain between $14.64\times$ and $16.60\times$. These checks confirm that the subset and scale phenomenon is not confined to one checkpoint, task, or model family.
\end{sloppypar}

\begin{table}[t!]
\centering
\caption{Robustness over 324 DistilBERT SST-2 audits. Means are over all six layers, Q--K and V--O pairs, subset sizes, and three shuffled seeds; square brackets are bootstrap 95\% confidence intervals for the mean here and in Tables~\ref{tab:cliprobust} and~\ref{tab:robustness_breadth}.}
\label{tab:robustness}
\small
\begin{tabular}{lccc}
\toprule
Gauge range & Raw top-3 overlap & Raw Spearman & Paired-product ratio \\
\midrule
Mild $[0.5,2.0]$ & $0.787$ $[0.741,0.833]$ & $0.844$ $[0.812,0.874]$ & $1.68$ $[1.66,1.69]$ \\
Medium $[0.25,4.0]$ & $0.605$ $[0.559,0.657]$ & $0.658$ $[0.614,0.701]$ & $4.43$ $[4.38,4.49]$ \\
Aggressive $[0.125,10.0]$ & $0.512$ $[0.466,0.559]$ & $0.550$ $[0.505,0.595]$ & $16.69$ $[16.42,16.95]$ \\
\bottomrule
\end{tabular}
\end{table}

The layer-level clipping audits of Table~\ref{tab:aggregate} use 48 examples per run. To test whether the effect persists with larger clipping subsets and multiple shuffles, Table~\ref{tab:cliprobust} reports a separate 216-audit clipping robustness suite over all six DistilBERT layers. The per-example gradient-norm ranking remains highly correlated across gauges: Under aggressive scaling, mean Spearman is $0.9978$, with individual runs ranging from $0.9911$ to $0.9998$. The absolute norm scale nevertheless shifts enough to change threshold decisions. Mean clipping flip rate rises smoothly from $3.38\%$ under mild gauges to $10.04\%$ under medium gauges and $18.97\%$ under aggressive gauges, with zero prediction flips in all runs. The 128-example subsets still show nonzero flip rates: $2.65\%$, $7.36\%$, and $15.32\%$ for mild, medium, and aggressive gauges, respectively. The clipping effect therefore persists at the larger audit size.

The 256-example holdout gives the same qualitative pattern on a seed and subset size not used in Table~\ref{tab:cliprobust}. Over layers 0, 3, and 5 and both Q--K and V--O pairs, prediction flips remain zero, and mean clipping flip rates are $2.34\%$, $7.16\%$, and $16.86\%$ for mild, medium, and aggressive gauges. The smaller means relative to the 48-example runs are expected because the empirical median threshold is less noisy at larger $n$, but the monotone scale dependence and nonzero mild-gauge flips persist.

\begin{table}[t!]
\centering
\caption{Per-example clipping robustness over 216 DistilBERT SST-2 audits. Means are aggregated as in Table~\ref{tab:robustness}, here over 48- and 128-example shuffled validation subsets.}
\label{tab:cliprobust}
\small
\begin{tabular}{lcccc}
\toprule
& \multicolumn{3}{c}{Clipping flip rate (\%)} & \\
\cmidrule(lr){2-4}
Gauge range & All subsets & $n=48$ & $n=128$ & Median-norm ratio \\
\midrule
Mild $[0.5,2.0]$ & $3.38$ $[2.88,3.91]$ & $4.11$ & $2.65$ & $1.20$ $[1.19,1.21]$ \\
Medium $[0.25,4.0]$ & $10.04$ $[8.74,11.41]$ & $12.73$ & $7.36$ & $1.86$ $[1.82,1.89]$ \\
Aggressive $[0.125,10.0]$ & $18.97$ $[17.46,20.65]$ & $22.63$ & $15.32$ & $3.50$ $[3.40,3.59]$ \\
\bottomrule
\end{tabular}
\end{table}

\begin{table}[t!]
\centering
\caption{Additional all-layer robustness checks. Each row aggregates 48 audits for a fixed gauge range: 12 layers, Q--K and V--O pairs, and two shuffled seeds.}
\label{tab:robustness_breadth}
\scriptsize
\begin{tabular}{llccc}
\toprule
Model and task & Gauge range & Raw top-3 overlap & Raw Spearman & Paired-product ratio \\
\midrule
\multirow{2}{*}{BERT on SST-2} & Mild $[0.5,2.0]$ & $0.701$ $[0.625,0.778]$ & $0.822$ $[0.775,0.866]$ & $1.67$ $[1.66,1.69]$ \\
 & Aggressive $[0.125,10.0]$ & $0.424$ $[0.347,0.507]$ & $0.478$ $[0.399,0.554]$ & $16.60$ $[16.23,16.99]$ \\
\addlinespace \multirow{2}{*}{ELECTRA on SST-2} & Mild $[0.5,2.0]$ & $0.736$ $[0.667,0.806]$ & $0.851$ $[0.809,0.889]$ & $1.66$ $[1.62,1.70]$ \\
 & Aggressive $[0.125,10.0]$ & $0.444$ $[0.375,0.514]$ & $0.488$ $[0.413,0.565]$ & $16.35$ $[15.27,17.60]$ \\
\addlinespace \multirow{2}{*}{RoBERTa on SST-2} & Mild $[0.5,2.0]$ & $0.625$ $[0.542,0.708]$ & $0.677$ $[0.581,0.764]$ & $1.61$ $[1.57,1.65]$ \\
 & Aggressive $[0.125,10.0]$ & $0.444$ $[0.361,0.528]$ & $0.366$ $[0.267,0.459]$ & $15.52$ $[14.58,16.47]$ \\
\addlinespace \multirow{2}{*}{RoBERTa on MNLI} & Mild $[0.5,2.0]$ & $0.556$ $[0.472,0.639]$ & $0.601$ $[0.493,0.700]$ & $1.59$ $[1.56,1.63]$ \\
 & Aggressive $[0.125,10.0]$ & $0.375$ $[0.306,0.444]$ & $0.323$ $[0.222,0.420]$ & $14.90$ $[14.12,15.71]$ \\
\addlinespace \multirow{2}{*}{RoBERTa on QNLI} & Mild $[0.5,2.0]$ & $0.597$ $[0.514,0.681]$ & $0.664$ $[0.558,0.759]$ & $1.58$ $[1.54,1.63]$ \\
 & Aggressive $[0.125,10.0]$ & $0.431$ $[0.361,0.500]$ & $0.354$ $[0.256,0.450]$ & $14.97$ $[14.03,16.05]$ \\
\addlinespace \multirow{2}{*}{RoBERTa on MRPC} & Mild $[0.5,2.0]$ & $0.583$ $[0.493,0.681]$ & $0.617$ $[0.510,0.723]$ & $1.60$ $[1.55,1.64]$ \\
 & Aggressive $[0.125,10.0]$ & $0.361$ $[0.285,0.438]$ & $0.309$ $[0.205,0.413]$ & $14.64$ $[13.73,15.65]$ \\
\bottomrule
\end{tabular}
\end{table}

\subsection{Checkpoint-Scale Norm Imbalance}

The aggressive gauges above are deliberately adversarial stress tests. To calibrate the gauge scale, we also measure the paired projection norms already present in the eight public encoder and ViT checkpoints in Table~\ref{tab:baseline}. Let $N_{\mathrm{left},h}$ and $N_{\mathrm{right},h}$ be the Euclidean norms of the complete branch parameter blocks that are actually rescaled: the relevant weight slice together with the corresponding Q, K, or V projection bias when present, while an unchanged output-projection bias is excluded. For each head, we report
\begin{equation}
    r_h\coloneqq\max\left\{ \frac{N_{\mathrm{left},h}}{N_{\mathrm{right},h}}, \frac{N_{\mathrm{right},h}}{N_{\mathrm{left},h}} \right\},
\end{equation}
where the two branches form either a Q--K pair or a V--O pair. The canonical scalar needed to equalize the pair is $s_h\coloneqq\sqrt{N_{\mathrm{right},h}/N_{\mathrm{left},h}}$.

The trained checkpoints already contain mild norm imbalance, especially in V--O pairs, but not the aggressive scales used for worst-case GRR stress tests. Across the eight checkpoints, the mean checkpoint-level Q--K median is $r_h=1.023$, the mean checkpoint-level 90th percentile is $1.063$, and the largest observed value is $1.187$; no Q--K head exceeds $1.25$, and the required balancing scales stay within $[0.918,1.055]$. For V--O, the corresponding checkpoint-summary means and largest observed value are $1.093$, $1.170$, and $1.429$; the checkpoint-level fraction above $1.25$ is $5.2\%$ on average and at most $10.4\%$, with balancing scales in $[0.837,1.077]$. Thus the mild robustness suite is closest to natural checkpoint-scale variation, while medium and aggressive gauges test whether a measurement is identifiable on the full same-function orbit.

This mild norm imbalance is consistent with the conservation-law perspective: For homogeneous factor pairs, gradient flow conserves the difference of squared branch norms, so training does not push a pair far apart when initialization uses small balanced weights \citep{du2018}. That argument constrains where optimization travels, not the orbit on which a measurement has to be identifiable.

We also fine-tune \path{distilbert-base-uncased} on the full SST-2 train split for 3 epochs with AdamW at learning rate $2\times10^{-5}$, tracking the same statistic throughout training. The reproducibility details give the complete recipe. Validation accuracy over epochs 0 through 3 is $50.92\%$, $90.37\%$, $90.71\%$, and $90.48\%$, respectively, ending at a level comparable to the public DistilBERT checkpoint used in the main audit. Across all 72 attention heads, however, the imbalance summary is unchanged to three decimal places: Q--K has median $1.015$ and maximum $1.187$, V--O has median $1.049$ and maximum $1.122$, and no head exceeds $1.25$. Paired norm imbalance therefore remains mild throughout standard fine-tuning.

\subsection{ViT Check}
\label{sec:vit-check}

We run the same attention-pair audit on ViT-base fine-tuned on CIFAR-10 to test whether the failure crosses modalities. The checkpoint obtains $98.51\%$ accuracy on the full 10,000-image test split, as recorded in Table~\ref{tab:baseline}. Across all 12 layers and both Q--K and V--O gauges, prediction flips remain zero. Raw head-gradient top-3 overlap ranges from $0$ to $1$ with median $1/3$. The corresponding mean ratio ranges from $2.59$ to $3.84$ with median $3.02$. Paired-product tangent overlap also ranges from $0$ to $1$ with median $1/3$. The tangent mean ratio ranges from $11.58$ to $27.22$ with median $17.91$. Per-example clipping flip rates range from $47.9\%$ to $50.0\%$ with median $50.0\%$.

The remedies behave as expected. Path-normalized gradients have top-3 overlap $1.000$ and mean ratio $1.00$, both with degenerate ranges at those values. Function-space head ablation and canonically balanced gradient scores likewise have top-3 overlap $1.000$ in all 24 audits and mean ratios approximately $1$. Raw top-3 ablation decisions, by contrast, have median Jaccard $0.2$ and yield an accuracy gap of $2.8$ percentage points on the full CIFAR-10 test split. The same symmetry-driven measurement failure therefore appears outside text encoders.

We also run a full-test global V--O pruning calibration on the same ViT checkpoint. The setup mirrors the RoBERTa global-profile audits: For each mild, medium, and aggressive same-function V--O profile, we compute one global raw-gradient ranking over all 144 heads using a 256-image audit subset, prune or ablate 22 heads, and evaluate the resulting mask on the full 10,000-image CIFAR-10 test split.

Table~\ref{tab:vit_global_profile} shows that the decision-level failure is not confined to text models. Mild gauges already change the top-ablation mask, with Jaccard $0.517$ and a full-test accuracy gap of $4.35$ percentage points. The aggressive profile gives Jaccard $0.189$ and a gap of $73.87$ percentage points: The original-coordinate top-head ablation yields $15.75\%$, while the reparameterized-coordinate ablation yields $89.62\%$. Bottom pruning is less destructive but still changes masks and produces a gap of up to $7.02$ percentage points. Canonical balancing restores identical global masks and zero accuracy gaps for all three profiles. Thus the ViT experiment provides both a layer-level measurement audit and a workflow-level pruning calibration.

\begin{table}[t!]
\centering
\caption{Full-test global V--O pruning calibration for ViT on CIFAR-10, using the same 98.51\% checkpoint and a 256-image gradient audit subset. Here and in later tables, pp denotes percentage points, and gap entries compare decisions from two same-function coordinates. Canonical balancing gives mask Jaccard $1.000$ and a $0.00$ pp accuracy gap in every row, so only raw results are shown.}
\label{tab:vit_global_profile}
\small
\begin{tabular}{llcc}
\toprule
Gauge profile & Decision & Raw mask Jaccard & Raw accuracy gap (pp) \\
\midrule
\multirow{2}{*}{Mild $[0.5,2.0]$} & bottom-15\% pruning & $0.913$ & $0.24$ \\
 & top-15\% ablation & $0.517$ & $4.35$ \\
\addlinespace \multirow{2}{*}{Medium $[0.25,4.0]$} & bottom-15\% pruning & $0.760$ & $4.97$ \\
 & top-15\% ablation & $0.189$ & $50.92$ \\
\addlinespace \multirow{2}{*}{Aggressive $[0.125,10.0]$} & bottom-15\% pruning & $0.517$ & $7.02$ \\
 & top-15\% ablation & $0.189$ & $73.87$ \\
\bottomrule
\end{tabular}
\end{table}

As a boundary control, we also test a quantity that should be a property of the fixed input--output function: the gradient of the supervised loss with respect to normalized input pixels. We apply the same global V--O profiles to all ViT attention layers and compare input-pixel gradients on 32 CIFAR-10 test images. Unlike parameter and internal-coordinate evidence, the input-pixel gradient is tied to the fixed input coordinate system. The control behaves accordingly: Across mild, medium, and aggressive profiles, the minimum per-image cosine similarity is $0.99998$, the largest relative $\ell_2$ difference is $0.79\%$, the mean norm ratio differs from $1$ by at most $1.01\times10^{-4}$, and prediction flips remain zero. The control delimits the claim precisely: The failure concerns parameter and internal gradient-derived measurements, not exact input-space derivatives of a preserved scalar function.

\subsection{Decoder-Only Language Modeling}
\label{sec:decoder-only}

We next audit decoder-only next-token prediction on WikiText-2. This decoder-only setting differs from GLUE in architecture, objective, output space, and decision metric. The loss is token-level cross-entropy, pruning is evaluated by perplexity, and the TinyLlama audit includes GQA with only four key and value heads shared by 32 query heads. Across all 24 layers of GPT-2-medium, all 36 layers of GPT-2-large, and all 22 layers of TinyLlama, with both Q--K and V--O gauges, the median next-token argmax flip rate is zero. GPT-2-medium and TinyLlama have zero flips in all audits. GPT-2-large has one finite-precision layer-9 Q--K case with flip rate $0.0492\%$ and maximum logit difference $9.54\times10^{-5}$.

Raw gradient evidence is again unstable, as Table~\ref{tab:gpt2} summarizes. GPT-2-medium raw head-gradient and paired-product tangent top-3 overlap both have median $1/3$, with zero-overlap failures. GPT-2-large shows the same medians over all 36 layers, with tangent overlap ranging from $0$ to $2/3$. TinyLlama has higher raw-gradient overlap, with median $1/2$, but the TinyLlama paired-product tangent overlap has median $0$; the tangent mean ratio has median $23.26\times$ and ranges from $12.37\times$ to $40.89\times$. Per-sequence clipping decisions flip for $50.0\%$ of audited sequences in every decoder-only run. Function-space head ablation remains invariant, with top-3 overlap $1.0$ in all 164 audits.

The layer-local pruning check is decision-facing as well. Raw bottom-3 head-pruning masks have Jaccard median $0.2$ for GPT-2-medium and GPT-2-large and median $0$ for TinyLlama. The induced WikiText-2 perplexity gap reaches $3.174$ points for GPT-2-medium on 16,256 tokens, $0.208$ points for GPT-2-large on 8,128 tokens, and $1.531$ points for TinyLlama on 8,128 tokens. Function-space bottom pruning selects the same heads in every decoder-only audit up to finite-precision ties and has negligible perplexity gaps. Thus the same reliability failure appears in generative decoder-only models, including a GQA Llama-style architecture. Figure~\ref{fig:summary}(b) plots the corresponding raw ranking and pruning-mask agreements for the three models.

\begin{figure}[t!]
\centering
\includegraphics[width=\linewidth]{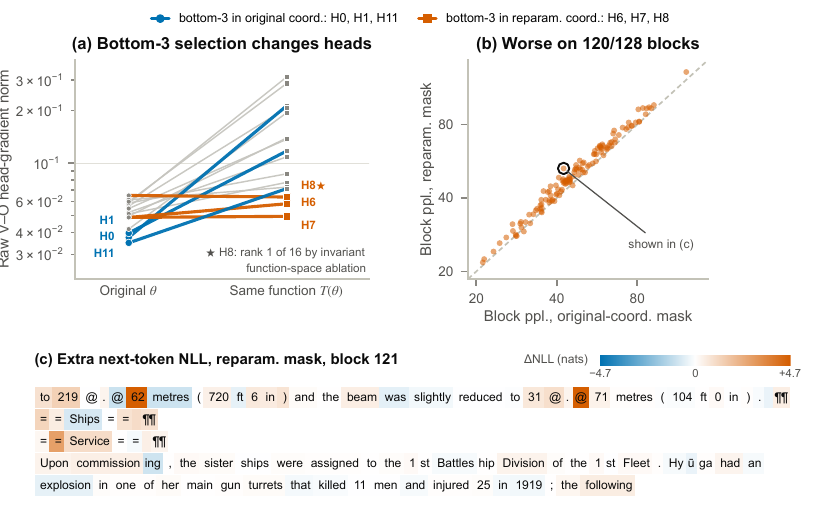}
\caption{Token-level anatomy of the largest layer-local decoder-only pruning gap: GPT-2-medium on WikiText-2, layer-2 V--O gauge, with raw-gradient bottom-3 pruning. Test perplexity is $46.75$ for the dense model, $46.92$ after pruning with the original-coordinate mask, and $50.09$ after pruning with the reparameterized-coordinate mask, giving a $3.17$-point gap. (a) With zero next-token argmax flips and maximum logit difference $3.36\times10^{-3}$, the coordinates select disjoint head sets with Jaccard $0$. (b) The reparameterized-coordinate mask is worse on 120 of 128 blocks. (c) The most affected block localizes the additional next-token negative log-likelihood. Canonical balancing makes the two raw-gradient masks identical, while function-space scores also agree across coordinates.}
\label{fig:qual-lm-tokens}
\end{figure}

Figure~\ref{fig:qual-lm-tokens} traces the largest GPT-2-medium layer-local gap from changed head identities to block- and token-level loss. The effect is broad across evaluation blocks rather than attributable to a single text fragment.

Global decoder-only pruning calibrations for GPT-2-medium and GPT-2-large are summarized in Table~\ref{tab:lm_global_profile}. The profile gauges preserve next-token argmax predictions on all audited validation tokens, with maximum logit drift below $4.0\times10^{-3}$ for GPT-2-medium and $1.2\times10^{-4}$ for GPT-2-large. Nevertheless, mild V--O scaling changes the global masks substantially. For GPT-2-medium, mild scaling over all 384 heads gives bottom-15\% Jaccard $0.634$, top-15\% Jaccard $0.487$, and an $8.68$ perplexity gap for top ablation. For GPT-2-large, mild scaling over all 720 heads gives bottom-15\% Jaccard $0.565$, top-15\% Jaccard $0.367$, and an $18.27$ perplexity gap for top ablation on the 8,128-token evaluation subset. In every profile and both model sizes, canonical V--O balancing makes the two same-function coordinates choose identical global masks and reduces the perplexity gap to numerical noise. For decoder-only language modeling, the failure therefore reaches a global perplexity-oriented pruning workflow and not only layer-local rankings.

\begin{table}[t!]
\centering
\caption{WikiText-2 decoder-only GRR checks over all layers. GPT-2-medium has 48 pair audits, GPT-2-large has 72, and TinyLlama has 44.}
\label{tab:gpt2}
\scriptsize
\resizebox{\linewidth}{!}{%
\begin{tabular}{llccc}
\toprule
Model & Measurement & Top-3 overlap or flip rate & Mean ratio & Decision effect \\
\midrule
\multirow{2}{*}{GPT-2-medium} & Raw head-gradient norm & $0.333$ $(0.000,1.000)$ & $2.61$ $(1.84,3.52)$ & pruning Jaccard med. $0.2$ \\
 & Raw paired-product tangent & $0.333$ $(0.000,1.000)$ & $13.84$ $(7.15,22.48)$ & max ppl. gap $3.174$ \\
\addlinespace \multirow{2}{*}{GPT-2-large} & Raw head-gradient norm & $0.333$ $(0.000,0.667)$ & $2.48$ $(2.05,3.62)$ & pruning Jaccard med. $0.2$ \\
 & Raw paired-product tangent & $0.333$ $(0.000,0.667)$ & $12.04$ $(5.29,22.66)$ & max ppl. gap $0.208$ \\
\addlinespace \multirow{2}{*}{TinyLlama 1.1B GQA} & Raw head-gradient norm & $0.500$ $(0.000,1.000)$ & $3.39$ $(2.16,5.19)$ & pruning Jaccard med. $0.0$ \\
 & Raw paired-product tangent & $0.000$ $(0.000,1.000)$ & $23.26$ $(12.37,40.89)$ & max ppl. gap $1.531$ \\
\midrule
\multirow{3}{*}{All three} & Per-sequence clipping & $50.0\%$ & Not applicable & Not applicable \\
 & Function-space ablation & $1.000$ $(1.000,1.000)$ & $\approx 1$ & stable \\
 & Canonically balanced gradient & $1.000$ $(1.000,1.000)$ & $\approx 1$ & stable \\
\bottomrule
\end{tabular}
}
\end{table}

\begin{table}[t!]
\centering
\caption{GPT-2-medium and GPT-2-large WikiText-2 global V--O pruning profiles. The 15\% pruning budget selects 58 of 384 heads for GPT-2-medium and 108 of 720 heads for GPT-2-large. Canonical balancing gives mask Jaccard $1.000$ and a perplexity gap below $10^{-4}$ in every row below, so only the raw columns are tabulated.}
\label{tab:lm_global_profile}
\small
\begin{tabular}{lllcc}
\toprule
Model & Gauge profile & Decision & Raw mask Jaccard & Raw ppl. gap \\
\midrule
\multirow{6}{*}{GPT-2-medium} & \multirow{2}{*}{Mild $[0.5,2.0]$} & bottom-15\% pruning & $0.634$ & $0.84$ \\
 & & top-15\% ablation & $0.487$ & $8.68$ \\
\addlinespace & \multirow{2}{*}{Medium $[0.25,4.0]$} & bottom-15\% pruning & $0.432$ & $0.52$ \\
 & & top-15\% ablation & $0.303$ & $1.71$ \\
\addlinespace & \multirow{2}{*}{Aggressive $[0.125,10.0]$} & bottom-15\% pruning & $0.349$ & $0.51$ \\
 & & top-15\% ablation & $0.234$ & $4.97$ \\
\midrule
\multirow{6}{*}{GPT-2-large} & \multirow{2}{*}{Mild $[0.5,2.0]$} & bottom-15\% pruning & $0.565$ & $0.04$ \\
 & & top-15\% ablation & $0.367$ & $18.27$ \\
\addlinespace & \multirow{2}{*}{Medium $[0.25,4.0]$} & bottom-15\% pruning & $0.479$ & $0.75$ \\
 & & top-15\% ablation & $0.220$ & $6.15$ \\
\addlinespace & \multirow{2}{*}{Aggressive $[0.125,10.0]$} & bottom-15\% pruning & $0.350$ & $0.77$ \\
 & & top-15\% ablation & $0.187$ & $7.52$ \\
\bottomrule
\end{tabular}
\end{table}

\subsection{Llama-Style SwiGLU MLP Channel Audit}
\label{sec:mlp-swiglu}

To test feed-forward channels, we audit TinyLlama SwiGLU up--down gauges at layers 0, 11, and 21, each with 5,632 channels. We use scales from $0.125$ to $10.0$, top-128 rankings on eight WikiText-2 blocks, and gradients restricted to each pair. Predictions never flip. Across layers, raw channel-gradient top-128 overlap ranges from $0.344$ to $0.422$ with median $0.352$. The mean ratio ranges from $2.45$ to $2.62$ with median $2.55$. Paired-product tangent overlap ranges from $0.211$ to $0.242$ with median $0.234$. The tangent mean ratio ranges from $9.84\times$ to $13.06\times$ with median $12.67\times$. Per-sequence clipping decisions flip for $50.0\%$ of audited sequences. Canonical balancing makes both overlaps and ratios $1.00$ throughout, showing that the failure extends beyond attention.

\subsection{Iterative Pruning Pipelines}
\label{sec:iterative-pruning}

One-shot pruning can understate the workflow consequence because practical pipelines often recompute scores after each masking step. We therefore run an iterative V--O schedule with RoBERTa on MRPC. Starting from original and reparameterized coordinates with identical validation predictions, each round recomputes global raw-gradient scores, selects seven additional heads, and accumulates the mask for four rounds. The final mask prunes 28 of 144 heads, or $19.4\%$. For the remedy check, we canonically balance both same-function coordinates before running the same iterative schedule.

\begin{figure}[t!]
\centering
\includegraphics[width=\linewidth]{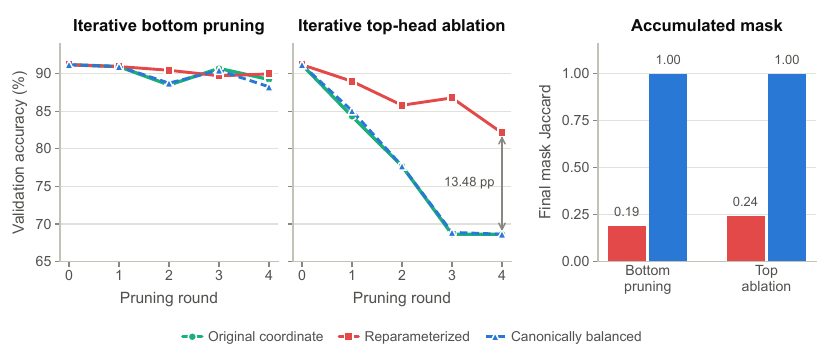}
\caption{Iterative pruning schedules for RoBERTa on MRPC. The right plot shows that raw same-function reparameterization changes the accumulated masks, and the center plot shows that the same reparameterization also changes the accuracy trajectory under top-head ablation. Canonical balancing maps both coordinates to the same schedule for the V--O scalar gauge, so the balanced curve tracks the original coordinate and the final mask Jaccard is $1.000$.}
\label{fig:iterative}
\end{figure}

Figure~\ref{fig:iterative} shows that same-function coordinate choice changes the entire schedule. Under bottom pruning, the final raw masks have Jaccard $0.191$ even though the accuracy curves remain close: The maximum gap is $1.96$ percentage points, and the final accuracies are $89.22\%$ and $89.95\%$ in the original and reparameterized coordinates, respectively. Under top-head ablation, the final raw masks have Jaccard $0.244$ and the curves separate: The largest validation gap is $18.14$ percentage points, and the final accuracies are $68.63\%$ and $82.11\%$ in the original and reparameterized coordinates, respectively, a gap of $13.48$ percentage points. Canonical balancing removes the schedule discrepancy for the V--O scalar gauge: Final Jaccard is $1.000$ for both decisions, with zero coordinate-induced accuracy gap and final accuracies of $88.24\%$ in both bottom-pruning runs and $68.63\%$ in both top-ablation runs. Thus GRR identifies instability in an iterative decision pipeline that repeatedly consumes gradient evidence.

\subsection{Normalization, Clipping, and Optimizer-State Audits}
\label{sec:adamw-state}

\paragraph{Fisher normalization.}

The targeted layer-11 V--O audit for RoBERTa on MRPC has zero forward prediction flips on the 128-example audit subset, yet the raw Euclidean head-gradient score again fails: Spearman correlation is negative, top-3 overlap is $1/3$, and the mean score ratio is $2.92$. Diagonal empirical-Fisher normalization is much more stable, with top-3 overlap $1.0$, Spearman $0.965$, and mean ratio $0.982$. This stability of the Fisher-normalized score is consistent with the expected cancellation of scalar-gauge factors in the normalized score. Gauge balancing gives a similar top-3 recovery for the V--O scalar gauge.

\paragraph{Clipped paired-product update.}
The same audit also measures a DP-SGD-style clipped update. Using the original median target-pair gradient norm as clipping threshold, $50.0\%$ of per-example clipping decisions flip after the same-function gauge. More importantly, the paired-product tangent induced by the clipped raw update is not just rescaled: The tangent's cosine similarity with the original-coordinate update is $0.425$, the relative $\ell_2$ difference is $10.90$, and the norm ratio is $11.29$. This comparison of clipped paired-product tangents links the clipping decision flips to an update-level consequence.

\paragraph{AdamW optimizer-state audit.}

The clipped-update result raises the related issue of whether consistent transport of AdamW moments across the same-function gauge makes the update a function-level object. We test the effect of transporting AdamW moments directly on the same layer-11 V--O pair of RoBERTa on MRPC. We accumulate AdamW first and second moments over 512 MRPC training examples without changing parameters. In the reparameterized coordinate, these directly accumulated moments match the original moments transported by the scalar gradient transformation: The relative $\ell_2$ errors are $4.48\times10^{-7}$ for the first moment and $2.94\times10^{-7}$ for the second. Thus the audit is not merely detecting a missing state transport.

Even with the transported AdamW moments, the paired-product tangent induced by one AdamW update is still coordinate-dependent. For both the directly accumulated transformed state and the transported original state, the update has cosine similarity $0.676$ with the original-coordinate product update, relative $\ell_2$ difference $2.934$, and norm ratio $3.515$. Canonical V--O balancing is much closer to the original update, with cosine similarity $0.995$, relative $\ell_2$ difference $0.107$, and norm ratio $1.013$. This optimizer-state audit separates two issues: Adam moments can be moved consistently as coordinate statistics, but the usual coordinate-wise AdamW step does not induce an invariant paired-product update under the audited gauge.

\subsection{Global Head-Pruning Case Study}
\label{sec:global-head-case-study}

\begin{figure}[ht!]
\centering
\includegraphics[width=\linewidth]{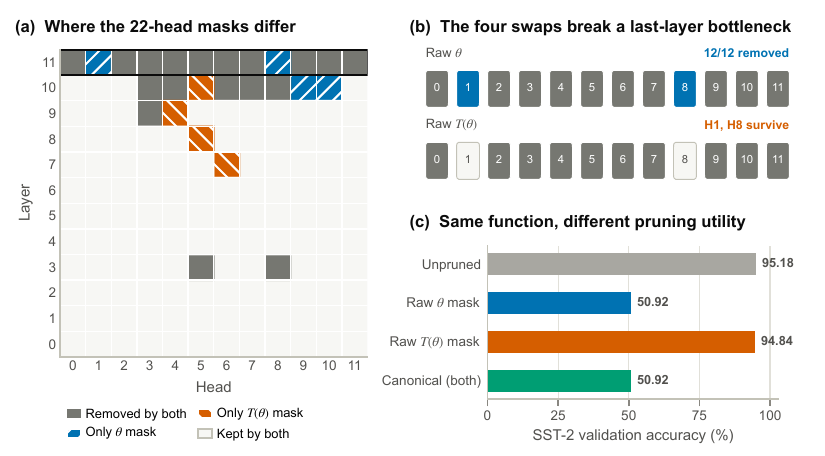}
\caption{Anatomy of the largest GLUE global-pruning gap: ELECTRA on SST-2 under an aggressive global Q--K gauge, with bottom-15\% pruning. (a) The two 22-head masks share 18 removed heads, with four removals unique to each coordinate. This difference amounts to four head substitutions and gives Jaccard $18/26=0.692$. (b) The original-coordinate mask removes every final-layer head, whereas the reparameterized-coordinate mask retains H1 and H8. Before any pruning, the two coordinates have zero prediction flips on the 128-example audit subset and a maximum logit difference of $1.25\times10^{-5}$. (c) The two masks yield $50.92\%$ and $94.84\%$ accuracy on the full 872-example validation set. Canonical balancing makes the masks identical, eliminating the coordinate discrepancy but not guaranteeing a useful pruning heuristic.}
\label{fig:qual-global-mask}
\end{figure}

Figure~\ref{fig:qual-global-mask} localizes the four head substitutions behind the largest GLUE global-pruning gap summarized in Table~\ref{tab:decisions}.

\subsection{Mild-Scale Downstream Calibration}

The largest global-pruning gaps reported so far use the aggressive head-scale profile. To make the scale issue explicit at the downstream-decision level, we run a profile-wise global V--O pruning calibration for RoBERTa on MRPC. We keep the same checkpoint, full 408-example validation split, 128-example audit subset, and global 15\% head budget, and change only the same-function gauge range.

Table~\ref{tab:global_profile} shows that mild gauges are sufficient to change the global-pruning decision. Under the mild $[0.5,2.0]$ V--O gauge, bottom-15\% raw-gradient pruning has mask Jaccard $0.630$ and a validation-accuracy gap of $17.89$ percentage points. The top-ablation mask also changes, with Jaccard $0.333$ and a gap of $7.35$ percentage points. As the gauge range increases, raw mask agreement decreases further, although the accuracy gap is not monotone because different selected head sets can have similar utility on MRPC. Canonical balancing removes the discrepancy for both decisions in all three profiles.

\begin{table}[t!]
\centering
\caption{Profile-wise global V--O pruning calibration for RoBERTa on MRPC. All rows use the same 91.18\% checkpoint and a 128-example gradient audit subset, and canonical balancing gives mask Jaccard $1.000$ and a $0.00$ pp accuracy gap throughout, so only the raw columns are tabulated.}
\label{tab:global_profile}
\small
\begin{tabular}{llcc}
\toprule
Gauge profile & Decision & Raw mask Jaccard & Raw accuracy gap (pp) \\
\midrule
\multirow{2}{*}{Mild $[0.5,2.0]$} & bottom-15\% pruning & $0.630$ & $17.89$ \\
 & top-15\% ablation & $0.333$ & $7.35$ \\
\addlinespace \multirow{2}{*}{Medium $[0.25,4.0]$} & bottom-15\% pruning & $0.517$ & $18.38$ \\
 & top-15\% ablation & $0.222$ & $3.68$ \\
\addlinespace \multirow{2}{*}{Aggressive $[0.125,10.0]$} & bottom-15\% pruning & $0.375$ & $19.85$ \\
 & top-15\% ablation & $0.158$ & $1.47$ \\
\bottomrule
\end{tabular}
\end{table}

\begin{figure}[t!]
\centering
\includegraphics[width=\linewidth]{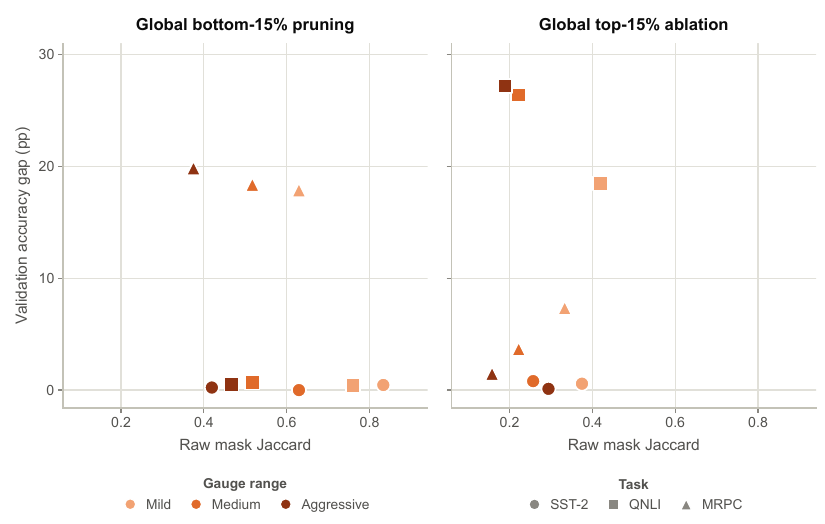}
\caption{Cross-task global V--O pruning calibration. Each point is a raw same-function comparison for a 15\% global-pruning decision with RoBERTa on SST-2, QNLI, or MRPC, with bottom pruning and top ablation split across the two plots. Mild gauges are enough to move masks substantially; whether that mask disagreement becomes a large validation gap is task- and decision-dependent.}
\label{fig:global_profiles}
\end{figure}

We then repeat the profile-wise global V--O calibration on RoBERTa checkpoints for SST-2 and QNLI, again using full validation splits for pruning accuracy. Table~\ref{tab:global_profile_tasks} shows that the effect is task-dependent but not MRPC-specific. On SST-2, mild gauges change the selected masks, but the full-validation utility gap remains below one percentage point. On QNLI, mild top-15\% ablation has raw mask Jaccard $0.419$ and a validation gap of $18.47$ percentage points; the aggressive profile reaches $27.18$ percentage points. Bottom pruning on QNLI has gaps below one percentage point despite changed masks. GRR therefore identifies whether the evidence used to choose a mask is a property of the learned function, independently of that mask's immediate utility. Figure~\ref{fig:global_profiles} plots every raw comparison in these three tasks.

\begin{table}[t!]
\centering
\caption{Cross-task profile-wise global V--O pruning calibration, using the gauge-profile family and audit budget of Table~\ref{tab:global_profile}. The maximum raw gap is taken over mild, medium, and aggressive profiles and both bottom-pruning and top-ablation decisions.}
\label{tab:global_profile_tasks}
\scriptsize
\begin{tabular}{lcccccc}
\toprule
& & \multicolumn{2}{c}{Mild bottom-15\% pruning} & \multicolumn{2}{c}{Mild top-15\% ablation} & \\
\cmidrule(lr){3-4}\cmidrule(lr){5-6}
Task and checkpoint & Baseline acc. (\%) & Mask Jaccard & Acc. gap (pp) & Mask Jaccard & Acc. gap (pp) & Max raw gap (pp) \\
\midrule
RoBERTa on SST-2 & $94.04$ & $0.833$ & $0.46$ & $0.375$ & $0.57$ & $0.80$ \\
RoBERTa on QNLI & $92.68$ & $0.760$ & $0.40$ & $0.419$ & $18.47$ & $27.18$ \\
RoBERTa on MRPC & $91.18$ & $0.630$ & $17.89$ & $0.333$ & $7.35$ & $19.85$ \\
\bottomrule
\end{tabular}
\end{table}

\subsection{Workflow Implications}
\label{app:workflow-implications}

\paragraph{Coordinate choices change pruning, clipping, and updates.}
Changing parameter coordinates affects four common uses of parameter gradients. In structured compression, raw scores can select different head and channel masks for the same unpruned model. In our largest GLUE global-pruning case, replacing four selected heads creates an accuracy gap of $43.92$ percentage points. The disagreement also persists through repeated decisions: After four rounds of top-head ablation on RoBERTa MRPC, the final masks have Jaccard $0.244$ and the accuracy gap is $13.48$ percentage points. In private training, a median $20.8\%$ of examples change clipping status under a same-function gauge. The privacy accounting is unchanged, but the set of clipped per-example gradients and the resulting clipping bias depend on the coordinates. Adapter workflows use factor gradients to allocate and prune rank budgets; the resulting rank-coordinate importance is basis-dependent unless defined through the product or a fixed gauge. Even a correctly transported AdamW state yields a coordinate-dependent paired-product update, so adaptive preconditioning alone does not remove the dependence. Each workflow therefore needs to check whether the pruning, clipping, rank-allocation, or update decisions remain unchanged when only the parameter coordinates change.

\section{Reproducibility Details}
\label{app:repro}

\subsection{Artifacts, Validation, and Traceability}

All reported numbers are generated by the Python audit drivers and serialized to a single result table. Each row is keyed by \texttt{experiment}, \texttt{quantity}, \texttt{method}, \texttt{transform}, and \texttt{metric}; the remaining columns are \texttt{value}, \texttt{baseline}, \texttt{transformed}, \texttt{forward\_max\_abs\_delta}, \texttt{n}, and \texttt{note}. A companion run-details record stores model tags, seeds, gauge scales, and selected heads or ranks. The full audit suite is executed in one pass by the suite runner under the full profile with all four GPUs visible and four concurrent shards. Separate entry points then regenerate the figure panels from the result table, export the table mapping, export the reproducibility manifest, validate the results, and print the reported aggregates.

The suite runner executes independent shards with one visible GPU per process and merges the per-shard result tables by replacing matching experiment names. Individual drivers also replace the rows and detail entries for the corresponding experiment family before appending new results, so rerunning one shard does not leave duplicate keys or discard unrelated completed audits. The merger checks for duplicate result keys, sorts rows by the result key, and atomically replaces the result and detail artifacts after all bytes have been written, making the reported tables insensitive to GPU completion order and interrupted writes. The runner writes one log per shard, treats a successful process that failed to write the expected result and detail artifacts as an error, and includes the tail of the shard log in failure messages.

The validation code checks finite numeric fields, unique result keys, baseline-quality thresholds, forward-preservation tolerances, required evidence rows for GLUE, ViT, decoder-only language modeling, global pruning, clipping robustness, AdamW optimizer-state auditing, training-time imbalance, and LoRA, representative reported values, and the expected figure and table-mapping artifacts. The baseline thresholds are set to reject weak checkpoints: For example, RoBERTa and ELECTRA on SST-2 must exceed $93\%$ and $94\%$ validation accuracy, respectively, ViT on CIFAR-10 must exceed $98\%$ test accuracy, and GPT-2 and TinyLlama WikiText-2 perplexities must remain below the stated limits. The artifact and hash checks are separate from the numeric checks, so result validation and reproducibility-artifact validation can be inspected independently.

\paragraph{Result-table traceability.}
The reported aggregates and figure panels are regenerated from the serialized result table. A declarative mapping associates every reported table or result family with the corresponding experiment filter, quantities, metrics, and representative value; an exporter then recomputes that mapping directly from the result table. The companion mapping artifact therefore provides the exact row-level trace from each reported result to the corresponding underlying records without requiring readers to search experiment logs.

We also export a reproducibility manifest. The manifest records SHA-256 hashes and sizes for the result table, the run-details record, the table mapping, and the generated figure PDFs. The manifest also records hashes for the experiment source files, package versions for PyTorch, Transformers, Datasets, PEFT, NumPy, pandas, SciPy, scikit-learn, Matplotlib, and Hugging Face Hub, CUDA and GPU metadata including the NVIDIA driver when \texttt{nvidia-smi} is available, the configured cache directories, git commit and dirty-state metadata, cached Hugging Face model snapshot hashes, and Hugging Face Datasets builder-cache fingerprints. For datasets, the manifest records the local cache directory, configuration, version, builder hash, \path{dataset_info.json} SHA-256 hash, and split shard sizes. Model and dataset revisions are not forced in the drivers. The manifest therefore records the cached local resource fingerprints to make the exact resources auditable. The validation script checks that the manifest hashes match the current result artifacts and experiment source files, so a result table or driver change cannot silently drift from the recorded reproducibility metadata.

\subsection{Data and Execution Configuration}

\paragraph{Dataset construction.}
GLUE validation accuracy and pruning accuracy use full validation splits for SST-2, QNLI, and MRPC and \texttt{validation\_matched} for MNLI. CIFAR-10 accuracy and pruning effects use the full public test split. WikiText-2 language-modeling perplexity uses contiguous 128-token blocks from the public test split. Gradient audits use deterministic prefixes or shuffled subsets with recorded seeds; robustness audits vary subset size and seed explicitly.

\paragraph{Random seeds.}
The main text reports how many seeds each suite uses; the exact values are the following. The 324-audit DistilBERT robustness suite and the 216-audit clipping robustness suite both shuffle the validation subsets with seeds 7, 11, and 13, and the all-layer 96-audit breadth checks use seeds 7 and 11. The 256-example clipping holdout uses seed 17, which no other robustness suite uses, so that suite is a genuine holdout in both subset size and seed. The six trained LoRA adapters use training seeds 7, 11, and 13 for both the rank-16 DistilBERT group on SST-2 and the rank-8 RoBERTa group on MRPC, and each LoRA rank-pruning recovery run inherits the seed of the source adapter. Audits that consume deterministic validation prefixes rather than shuffled subsets, including the main GLUE layer sweep, the ViT audits, and the decoder-only WikiText-2 audits, do not depend on a seed choice. The seed of every recorded run is stored together with the model tag and gauge scales in the run-details record.

\paragraph{Training recipes.}
The main classifiers and ViT are public fine-tuned checkpoints. The LoRA experiments train DistilBERT-base on SST-2 for 3 epochs with rank 16, $\alpha=32$, dropout 0.05, AdamW, learning rate $2\times10^{-4}$, weight decay 0.01, linear warmup ratio 0.06, batch size 64, and gradient clipping at 1.0. The RoBERTa LoRA experiments on MRPC start from \path{textattack/roberta-base-MRPC} and train rank-8 adapters for 3 epochs with $\alpha=16$, AdamW, learning rate $10^{-4}$, weight decay 0.01, linear warmup ratio 0.06, batch size 32, Q, K, and V target modules, and the classifier saved with the adapter.

LoRA rank-pruning recovery keeps the selected rank coordinates zero and fine-tunes the remaining trainable adapter and classifier parameters for 2 epochs: SST-2 uses learning rate $10^{-4}$ and batch size 64, while MRPC uses learning rate $5\times10^{-5}$ and batch size 32. The training-time imbalance calibration fine-tunes all parameters of \path{distilbert-base-uncased} on the full SST-2 train split for 3 epochs with AdamW, learning rate $2\times10^{-5}$, weight decay 0.01, linear warmup ratio 0.06, batch size 64, and gradient clipping at 1.0; validation accuracy reaches $90.71\%$ after epoch 2.

\paragraph{Optimizer and pruning audit settings.}
The AdamW optimizer-state audit uses the same layer-11 V--O pair of RoBERTa on MRPC, accumulates $\beta_1=0.9,\beta_2=0.999$ moments over 512 MRPC train examples in 16-example batches without optimizer steps, and compares the one-step AdamW paired-product tangent with $\epsilon=10^{-8}$ and weight decay 0.01. The iterative pruning audit uses the RoBERTa checkpoint on MRPC, a 128-example validation audit subset for gradient recomputation, the full 408-example validation split for accuracy, four rounds, and seven newly selected global V--O heads per round.

\paragraph{Global-pruning and control settings.}
The GLUE profile-wise global-pruning calibrations use RoBERTa checkpoints on SST-2, QNLI, and MRPC, full validation splits, a 128-example audit subset, global V--O decisions for bottom-15\% pruning and top-15\% ablation, and mild, medium, and aggressive gauge profiles. The ViT global-pruning calibration uses \path{nateraw/vit-base-patch16-224-cifar10}, a 256-image test audit subset for gradients, the full 10,000-image CIFAR-10 test split for accuracy, the same 15\% global V--O head budget, and the same three gauge profiles.

The ViT input-gradient control uses the same checkpoint, 32 CIFAR-10 test images, global V--O gauges at mild, medium, and aggressive scales, and gradients of the supervised loss with respect to normalized input pixels. The GPT-2-medium global-pruning calibration uses all 384 attention heads, 32 WikiText-2 validation audit blocks for gradients, 128 WikiText-2 test blocks for perplexity, the same 15\% global V--O head budget, and the same three gauge profiles. The SwiGLU MLP audit uses the TinyLlama checkpoint, layers 0, 11, and 21, eight 128-token validation blocks for gradient scoring, eight one-sequence blocks for clipping decisions, top-128 channel rankings, and gradients restricted to the audited up--down projection pair.

The GPT-2-large global V--O profile reported in Table~\ref{tab:lm_global_profile} uses all 720 attention heads, 16 WikiText-2 validation audit blocks for gradients, 64 WikiText-2 test blocks for perplexity, batch size 2, the same 15\% global head budget, and the same mild, medium, and aggressive gauge profiles as the GPT-2-medium calibration.

\paragraph{Hardware and precision.}
The reported run was executed on a 4$\times$NVIDIA H100 node with about 96GB memory per GPU. The suite runner assigns one shard process per visible GPU and uses four concurrent GPU-isolated shards. The largest single-GPU shards are the TinyLlama attention and MLP audits and the GPT-2-large global pruning profile; the configured block counts and batch sizes fit on one H100. With warm Hugging Face caches, the full suite completes as an overnight multi-GPU job. For forward-preservation checks, TF32 matrix multiplication is disabled and FP32 matmuls are used to avoid conflating finite-precision kernel differences with symmetry failure.

\section{A GRR Checklist for New Gradient-Based Diagnostics}
\label{app:checklist}

The following procedure produces the GRR report of Definition~\ref{def:grr} for a diagnostic that was not studied here. Appendix~\ref{app:minimal-grr-code} gives an executable version for the attention case, and the settings quoted below match the settings used throughout our experiments.

\paragraph{Applying GRR.}
To apply GRR, first identify a parameter transformation that preserves the model's input--output function exactly. Apply the transformation at several scales, from mild rescaling to a stronger stress test, and verify that logits and predictions remain unchanged within numerical tolerance. Reject runs that fail this forward check before comparing gradients. Then compare both the gradient-based scores and the pruning, clipping, or rank-selection decisions those scores produce. Repeat the comparison with a control that is invariant to the tested transformation, using a function-space score for the same intervention or a quotient-space score for a bilinear product. If the function is unchanged but the workflow selects different heads, examples, or rank coordinates, the decision depends on the parameter representation and should not be presented as a property of the learned function. The checklist below gives the recommended gauge profiles and acceptance criteria.

\paragraph{Exact symmetry of the audited architecture.}
Standard multi-head attention admits head-wise Q--K and V--O scalar gauges, SwiGLU blocks admit up--down channel gauges, and low-rank adapters admit the full $GL(r)$ factor gauge. Under RoPE, only the scalar part of the Q--K gauge survives; under GQA, the gauge acts on shared key--value groups rather than on individual query heads. Verify exactness symbolically before running anything, since an approximate symmetry converts every later comparison into a confound. If the audited component admits no exact symmetry, report the absence of an exact symmetry as an informative result about identifiability and stop.

\paragraph{Gauge profiles.}
We recommend three gauge profiles. A mild profile spanning $[0.5,2.0]$ brackets the paired-norm imbalance already present in public checkpoints, which the checkpoint-scale audit in Appendix~\ref{app:additional-experiments} places within balancing scales $[0.837,1.077]$. A medium profile spans $[0.25,4.0]$. An aggressive profile spanning $[0.125,10.0]$ probes identifiability on the full orbit. Reporting only the aggressive profile invites the objection that the gauge is artificial, whereas reporting only the mild profile understates how badly the measurement is underdetermined.

\paragraph{Forward preservation.}
Report the maximum logit difference $\delta_f(T)$ and the prediction flip rate on the audit set. Discard any run whose flips are not attributable to finite precision; in our audits the classification flip rate is zero throughout, and the only nonzero decoder-only case is a finite-precision GPT-2-large flip rate of $0.0492\%$. A run that fails the forward-preservation gate describes a broken transformation rather than an unreliable diagnostic, and reporting such a run as a GRR failure is an error.

\paragraph{Measurement stability.}
Report the rank correlation $\rho_R$, the top-$k$ overlap $O_k$ at the $k$ that the workflow actually uses, and the ratio of mean scores. A mean ratio far from one indicates that any absolute threshold placed on the score, such as a clipping threshold or a pruning cutoff, is gauge-dependent even when the induced ordering happens to survive.

\paragraph{Decision stability.}
This decision-level comparison is the step that carries the claim. Report the agreement of the selected sets, for instance the Jaccard index of two pruning masks, together with the gap in the metric the workflow optimizes. A stable score with an unstable decision and a stable decision with an unstable score are both possible, and only the decision-level comparison states what is at stake for a practitioner. Our SST-2 profiles illustrate the second case: Masks change while the utility gap stays below one percentage point, so evidence instability and immediate utility loss are distinct findings and should be reported separately.

\paragraph{Invariant controls.}
Function-space ablation is the most direct control and the most expensive, since the ablation requires a forward evaluation per candidate component. The path-normalized score is nearly free and cancels reciprocal scalar gauges exactly. Canonical balancing fixes the coordinate before measuring and therefore lets an existing diagnostic be reused unchanged. None of these three controls addresses non-diagonal $GL(r)$ mixing; when the audited gauge includes a basis rotation, the control has to be defined on the quotient, as in Corollary~\ref{cor:quotient-gradient}.

\paragraph{Reading the outcome.}
If the decision is stable across all three profiles, the diagnostic may be reported as evidence about the learned function under the audited symmetry. If the decision changes under the aggressive profile only, the diagnostic is not identifiable on the full orbit but may still be usable when the coordinate is pinned by an explicit convention, which should then be stated wherever the score is reported. If the decision changes under the mild profile, the diagnostic should be canonicalized before use or replaced by a quotient-space score, because coordinate variation of the size already present in ordinary checkpoints is enough to change the conclusion.

\section{Limitations}
\label{app:limitations}

The main limitation of this work is the scope of the empirical evaluation. The audits cover encoder classifiers on four GLUE tasks, a ViT on CIFAR-10, three decoder-only language models and a SwiGLU MLP channel audit on WikiText-2, and six trained LoRA adapters, which is a broad but still finite selection rather than a sweep over all datasets, model families, and model scales. Coverage is also uneven across workflows: The vision, decoder-only, and MLP audits use fewer model--task combinations than the GLUE sweep, and the iterative pruning curves are centered on RoBERTa and MRPC. Because the audited symmetries are exact and architecture-native rather than checkpoint-specific, the same reliability concern applies to settings we did not run, and extending the audit to further benchmarks, larger models, and additional gradient-based workflows is direct future work.

\end{document}